%% file: PaperForReview.tex
\documentclass[10pt,twocolumn,letterpaper]{article}

\usepackage{cvpr}              

\usepackage{graphicx}
\usepackage{amsmath}
\usepackage{amssymb}
\usepackage{booktabs}
\usepackage{multirow}
\usepackage{color}

\usepackage[table]{xcolor}
\usepackage{makecell}

\usepackage[pagebackref,breaklinks,colorlinks]{hyperref}

\usepackage[capitalize]{cleveref}
\crefname{section}{Sec.}{Secs.}
\Crefname{section}{Section}{Sections}
\Crefname{table}{Table}{Tables}
\crefname{table}{Tab.}{Tabs.}

\def\cvprPaperID{5710} 
\def\confName{CVPR}
\def\confYear{2023}

\begin{document}

\title{HumanBench: Towards General Human-centric Perception \\ with Projector Assisted Pretraining}

\author{Shixiang Tang\textsuperscript{1,4}\thanks{Equal contribution. This work was done in SenseTime.}, Cheng Chen\textsuperscript{4}\footnotemark[\value{footnote}], Qingsong Xie\textsuperscript{4}, Meilin Chen\textsuperscript{2,4}, Yizhou Wang\textsuperscript{2,4}, Yuanzheng Ci\textsuperscript{1},\\
Lei Bai\textsuperscript{3}\thanks{Corresponding author.},
Feng Zhu\textsuperscript{4}, Haiyang Yang\textsuperscript{4}, Li Yi\textsuperscript{4}, Rui Zhao\textsuperscript{4,5}, Wanli Ouyang\textsuperscript{3} \\
{\textsuperscript{1}The University of Sydney, \textsuperscript{2}Zhejiang University, \textsuperscript{3}Shanghai AI Laboratory, \textsuperscript{4}SenseTime Research},\\ {\textsuperscript{5}Qing Yuan Research Institute, Shanghai Jiao Tong University, China} \\
\tt\small stan3906@uni.sydney.edu.au, chencheng1@sensetime.com, bailei@pjlab.org.cn
}

\maketitle

\begin{abstract}
Human-centric perceptions include a variety of vision tasks, which have widespread industrial applications, including surveillance, autonomous driving, and the metaverse.  It is desirable to have a general pretrain model for versatile human-centric downstream tasks. This paper  forges ahead along this path from the aspects of both benchmark and pretraining methods. Specifically, we propose a \textbf{HumanBench} based on existing datasets to comprehensively evaluate on the common ground the generalization abilities of different pretraining methods on 19 datasets from 6 diverse downstream tasks, including person ReID, pose estimation, human parsing, pedestrian attribute recognition, pedestrian detection, and crowd counting. To learn both coarse-grained and fine-grained knowledge in human bodies, we further propose a \textbf{P}rojector \textbf{A}ssis\textbf{T}ed \textbf{H}ierarchical pretraining method (\textbf{PATH}) to learn diverse knowledge at different granularity levels. Comprehensive evaluations on HumanBench show that our PATH achieves new state-of-the-art results on 17 downstream datasets and on-par results on the other 2 datasets. The code will be publicly at \href{https://github.com/OpenGVLab/HumanBench}{https://github.com/OpenGVLab/HumanBench}.
\end{abstract}

\section{Introduction}
\label{sec:intro}


\begin{figure*}[t]
    \centering
    \includegraphics[width=0.95\linewidth]{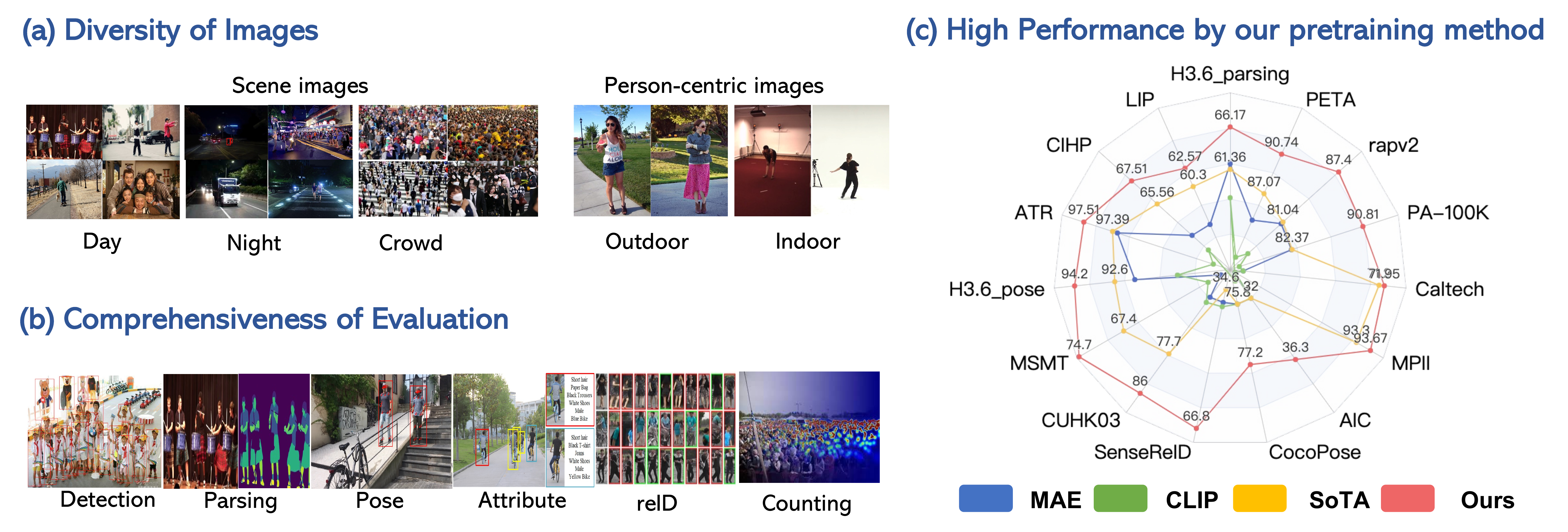}
    \caption{(a-b) Overview of our proposed HumanBench. HumanBench includes diverse images, including scene images and person-centric images. Our HumanBench also has comprehensive evaluation. Specifically, it evaluates pretraining models on 6 tasks, including pedestrian detection, human parsing, pose estimation, pedestrian attribute recognition, person ReID, and crowd counting. (c) High performances are achieved by our pretraining method on HumanBench. We report 1-heavy occluded MR$^{-2}$ and 1-EPE for Caltech and H3.6pose.}
    \vspace{-1.5em}
    \label{fig:teaser}
\end{figure*}

Human-centric perception has been a long-standing pursuit for computer vision and machine learning communities. It encompasses massive research tasks and applications including person ReID in surveillance \cite{zheng2021online,ge2020self,luo2019bag,ge2020mutual,yu2021multiple}, human parsing and pose estimation in the metaverse \cite{xu2022vitpose, tu2020voxelpose,xiao2018simple,li2021tokenpose,li2022mvitv2,ma2017pose}, and pedestrian detection in autonomous driving~\cite{chu2020detection,lin2020detr,wang2022anchor}. 
Although significant progress has been made, most existing human-centric studies and pipelines are task-specific for better performances, leading to huge costs in representation/network design, pretraining, parameter-tuning, and annotations.
To promote real-world deployment, we ask: \emph{whether a general human-centric pretraining model can be developed that can benefit diverse human-centric tasks and be efficiently adapted to downstream tasks?}


Intuitively, we argue that pretraining such general human-centric models is possible for two reasons. First, there are obvious correlations among different human-centric tasks. For example, both human parsing and pose estimation predict the fine-grained parts of human bodies~\cite{liang2018look,hong2022versatile} with differences in annotation granularities. Thus, the annotations in one human-centric task may benefit other human-centric tasks when trained together. 
Second, recent achievements in foundation models~\cite{vaswani2017attention, devlin2018bert, radford2018improving, radford2019language, brown2020language, lepikhin2020gshard} have shown that large-scale deep neural networks (\emph{e.g.}, transformers \cite{dosovitskiy2020image}) have the flexibility to handle diverse input modalities and the capacity to deal with different tasks. For example, Uni-Percevier~\cite{zhu2022uni} and BEITv3~\cite{wang2022image} are applicable to multiple vision and language tasks.  

Despite the opportunities of processing multiple human-centric tasks with one pretraining model, there are two obstacles for developing general human-centric pretraining models. First, although there are many benchmarks for every single human-centric task, there is still no benchmark to fairly and comprehensively compare various pretraining methods on a common ground for a broad range of human-centric tasks, data distributions, and application scenarios. Second, different from most existing general foundation models trained by unified global vision-language consistencies, pretraining human-centric models are required to learn both global (\emph{e.g.,} person ReID and pedestrian detection) and fine-grained semantic features (\emph{e.g.,} pose estimation and human parsing) of human bodies from diverse annotation granularity simultaneously.

In this paper, we first build a benchmark, called \textbf{HumanBench}, based on existing datasets to enable pretraining and evaluating human-centric representations that can be generalized to various downstream tasks.  
HumanBench has two appealing properties. \textbf{(1) Diversity.} The images in our HumanBench include diverse image properties, ranging from person-centric cropped images to scene images with crowd pedestrians, ranging from indoor scenes to outdoor scenes (Fig.~\ref{fig:teaser}(a)), and from surveillance to metaverse. \textbf{(2) Comprehensiveness.} Humanbench covers comprehensive image-based human-centric tasks in both pretraining datasets and downstream tasks (Fig.~\ref{fig:teaser}(b)). For pretraining, we include 11 million images from 37 datasets across five representative human-centric tasks, \emph{i.e.,} person ReID, pose estimation, human parsing, pedestrian attribute recognition, and pedestrian detection. For evaluation, HumanBench evaluates the generalization abilities on 12 pretraining datasets, 6 unseen datasets of pretraining tasks, and 2 datasets out of pretraining tasks, ranging from global prediction, \emph{i.e.,} ReID, to local prediction, \emph{i.e.,} human parsing and pose estimation. Results on our HumanBench (Fig.~\ref{fig:teaser}(c)) 
lead to two interesting findings. First, compared with datasets with natural images for general pretrained models,  HumanBench is more effective for human-centric perception tasks. Second, as human-centric pretraining requires to learn features of diverse granularity, supervised pretraining methods with proper designs can learn from diverse annotations in HumanBench and perform better than the existing unsupervised pretraining methods, for which details will be shown in Sec.~\ref{sec:ablation}.

Based on HumanBench, we further investigate how to learn a better human-centric supervised pretraining model from diverse datasets with various annotations. However, naive multitask pretraining may easily suffer from the task conflicts~\cite{liu2021conflict,yu2020gradient} or overfitting to pretrained annotations~\cite{sariyildiz2021concept,zhao2020makes}, losing the desirable generalization ability of pretraining. Inspired by~\cite{wang2022revisiting}, which suggests adding an MLP projector before the task head can significantly enhance the generalization ability of supervised pretraining, we propose \underline{\textbf{P}}rojector \underline{\textbf{A}}ssis\underline{\textbf{T}}ed \underline{\textbf{H}}ierarchical Pre-training (\textbf{PATH}), a projector assisted pretraining method with hierarchical weight sharing to tackle the task conflicts of supervised pretraining from diverse annotations. Specifically, the weights of backbones are shared among all datasets, and the weights of projectors are shared only for datasets of the same tasks, while the weights of the heads are shared only for a single dataset -- forming a hierarchical weight-sharing structure. 
During the pretraining stage, we insert the task-specific projectors before dataset heads but discard them when evaluating models on downstream tasks. With the hierarchical weight-sharing strategy, our pretraining method enforces the backbone to learn the shared knowledge pool, the projector to attend to the task-specific knowledge, and the head to focus on the dataset with specific annotation and data distribution.

In summary, our contributions are two folds: (1) we build HumanBench, a large-scale dataset for human-centric pretraining including diverse images and comprehensive evaluations. (2) To tackle the diversity of input images and annotations of various human-centric datasets, we propose PATH, a projector-assisted hierarchical weight-sharing method for pretraining the general human-centric representations. We achieve state-of-the-art results by PATH on 15 datasets throughout 6 downstream tasks (Fig.~\ref{fig:teaser}(c)), on-par results on 2 datasets, and slightly lower results on 2 datasets on HumanBench when using ViT-Base. Experiments with ViT-Large backbone show that our method can further achieve considerable gains over ViT-Base, achieving another 2 new state-of-the-art results and showing the promising scalability of our method. We hope our work can shed light on future research on pretraining human-centric representations, such as unified structures.



\section{Related Work}
\noindent \textbf{General Vision Models.}
The vision community has witnessed the emergence of general vision models~\cite{yuan2021florence, shao2021intern,wang2022image,zhu2022uni}, which could learn complex patterns from large-scale data and are powerful when adapted for downstream tasks. Although they achieve promising results, they pretrain on natural images, which are sub-optional for human-centric perception tasks. Furthermore, most of them leverage contrastive learning or masked reconstruction on global images, but do not learn multi-scale representations of human bodies with a diverse granularity that is important for diverse human-centric perceptions. In this paper, we tackle this problem by supervised pretraining from large-scale publicly available datasets with released annotations. The massive human body annotations with diverse granularity enable the model to learn multi-scale features, which is preferred in various human-centric tasks. SL-MLP~\cite{wang2022revisiting} is an improved supervised pretraining method that proposes to insert an MLP projector to increase the generalization ability of pretraining models. However, it does not tackle the task conflict problem in a multitask pretraining as our method.




\noindent \textbf{Pretraining on Human-centric Tasks.}
Human-centric perception has been studied for decades. However, there are only a few works on pretraining diverse human-centric tasks. HCMoCo~\cite{hong2022versatile} pretrains human representations for only pose estimation and human parsing with human-centric images. Our work pretrains a model for more diverse human-centric tasks, ranging from global identification to local prediction, from tasks on cropped human-centric images to scene images. Furthermore, different from HCMoCo which mainly learns human-centric representation from the multi-modal (RGB-D) representation of the same image, we learn human representations from multi-datasets where RGB images with one or two kinds of annotation, which is a more practical and data-scalable setting for pretraining. For example, publicly available datasets for ReID, attribute, and counting do not have RGB-D data.

\begin{table}[t]
  \centering
  \footnotesize
  \caption{Statistics of Pretraining Datasets}
    \begin{tabular}{ccc}
    \toprule
    Task  & Number of datasets & Number of Images \\
    \midrule
    Person ReID  & 7     & 5,446,419 \\
    Pose estimation  & 11    & 3,739,291 \\
    Human parsing & 7     & 1,419,910 \\
    Pedestrian Attribute & 6     & 242,880 \\
    Pedestrian Detection & 6     & 170,687 \\
    \midrule
    In total & 37    & 11,019,187 \\
    \bottomrule
    \end{tabular}%
  \label{tab:pretraining}%
  \vspace{-1em}
\end{table}%

\noindent \textbf{Multi-task  Multi-dataset  Pretraining.}
Multi-task and Multi-dataset Pretraing is a popular framework for pretraining models~\cite{zhu2022uni,wang2022ofa,he2022x}. However, they may easily suffer from task conflicts. There are generally two routes to reduce task conflicts, including network designs~\cite{he2017mask, liu2019end, misra2016cross, teichmann2018multinet, kokkinos2017ubernet, lu2017fully} and optimization strategies~\cite{kendall2018multi, chen2018gradnorm, liu2019end, guo2018dynamic, sener2018multi}. Different from existing methods to tackle task conflicts from the balance of losses, Our PATH explores a new network optimization strategy from framework design at the training stage. Specifically, we insert a task-specific projector into the backbone and the dataset head, and design the parameters in the backbone, the parameters in the task-specific projector and the head are shared among all datasets, datasets of the same task and not shared. With this strategy, the dataset-specific heads and the task-specific projectors are the regularizations for the backbone of learning shared knowledge across various human-centric tasks. Our method does not design new networks because we only evaluate the generalization of the backbone.

\section{HumanBench}

%
\subsection{Pretraining Datasets} \label{sec:pretrain_task}
According to biologists~\cite{darwin1998expression}, nonverbal communication in daily life includes identity, visual appearance, and posture information. Following this domain knowledge, we select person ReID as the identification task, pedestrian attribute recognition, pedestrian detection, human parsing as the visual appearance task, and pose estimation as the posture task in HumanBench. 37 datasets containing 11,019,187 images \footnote{Full list of these 37 datasets are given in the supplementary.} are collected  for pretraining. Tab.~\ref{tab:pretraining} presents the number of datasets and images in each task. For the selected datasets, we leverage their original annotations except for  the noisy labeled person ReID dataset, \emph{i.e.,} LUPerson-NL. In LUPerson-NL, we observe that identities with relatively few images are accurate. Therefore, we only select the identities that contain 15 to 200 images in LUPerson-NL, corresponding to 151,595 identities and 5,178,420 images. 

To ensure no data leakage and small information redundancy, we further de-duplicate the pretraining dataset from two aspects.
First, we remove all potential duplicates from pretraining datasets that may appear in the evaluation datasets (detailed in Sec.~\ref{sec:evaluation_protocol}) to enable a meaningful evaluation of generalization. Specifically, we first utilize the Difference Hash~\cite{suryawanshi2018image} to calculate the hash code of images in the evaluation datasets and pretraining datasets. Then, we delete the images in the pretraining datasets that have the same hash code as any image in the evaluation datasets. Second, some images come from some video-based datasets, \emph{e.g.,} AIST++~\cite{li2021ai} and UppenAction~\cite{zhang2013actemes},  which contain large information redundancy between consecutive frames. In this case, we select only one image from every 8 consecutive frames to reduce redundancy.

\begin{table}[t]
  \centering
  \caption{Summary of datasets for in-dataset evaluations, out-of-dataset evaluations, and unseen-task evaluations.}
  \resizebox{0.45\textwidth}{!}{
    \begin{tabular}{cc|cp{2cm}cp{2cm}cp{2cm}}
    \toprule
    Task  & Datasets & \makecell[c]{in-dataset\\ evaluations} & \makecell[c]{out-of-dataset \\ evaluations}  &  \makecell[c]{unseen-task\\ evaluations} \\
    \midrule
    \multirow{4}[4]{*}{ReID} & Market1501~\cite{zheng2015scalable} & \checkmark & & \\
          & MSMT~\cite{wei2018person}  &    \checkmark   &  & \\
          & CUHK03~\cite{li2014deepreid} &   \checkmark    & & \\
\cmidrule{2-5}          & SenseReID~\cite{zhao2017spindle} & &\checkmark    &  \\
    \midrule
    \multirow{4}[4]{*}{Pose} & COCO~\cite{lin2014microsoft}  & \checkmark & & \\
          & Human3.6M~\cite{h36m_pami} &   \checkmark    &  & \\
          & AIC~\cite{wu2019large}   &    \checkmark   &  & \\
\cmidrule{2-5}          & MPII~\cite{andriluka20142d}  & & \checkmark     &  \\
    \midrule
    \multirow{4}[4]{*}{Parsing} & Human3.6M~\cite{h36m_pami} & \checkmark & & \\
          & LIP~\cite{gong2017look}   &   \checkmark    &  & \\
          & CIHP~\cite{gong2018instance}  &   \checkmark    &   & \\
\cmidrule{2-5}          & ATR~\cite{liang2015human}   & & \checkmark     &  \\
    \midrule
    \multirow{3}[4]{*}{Attribute} & PA-100K~\cite{liu2017hydraplus} & \checkmark & & \\
          & RAPv2~\cite{li2018richly} &    \checkmark   &  & \\
\cmidrule{2-5}          & PETA~\cite{deng2014pedestrian} & & \checkmark     &  \\
    \midrule
    \multirow{2}[2]{*}{Detecton} & CrowdHuman~\cite{shao2018crowdhuman} & \checkmark     & & \\
    \cmidrule{2-5}  & Caltech~\cite{dollar2011pedestrian} & & \checkmark     &  \\
    \midrule
    \multirow{2}[2]{*}{Counting} & ShTech PartA~\cite{zhang2016single} &  & &\checkmark \\
          & ShTech PartB~\cite{zhang2016single} &       & &\checkmark \\
    \bottomrule
    \end{tabular}
    }
  \label{tab:downstream}%
\end{table}%




\subsection{Evaluation Scenarios and Protocols} \label{sec:evaluation_protocol}

\noindent \textbf{Evaluation Scenarios.}
Our benchmark comprehensively quantifies the generalization ability of human-centric representation on 6 human-centric tasks from 19 datasets. Specifically, we establish three evaluation scenarios for HumanBench: (1) \emph{in-dataset evaluation:} we select 12 representative datasets whose training subsets are allocated to the pretraining dataset and evaluation subsets assigned to the evaluation dataset to evaluate the performance of a general pretrained model on diverse seen datasets (meaning similar data distribution for training and evaluation). (2) \emph{out-of-dataset evaluation:} we select 5 datasets that do not appear in pretraining but belong to the seen task for evaluating the ability of the pretrained model on unseen datasets (meaning potentially different data distribution for training and evaluation). (3) \emph{unseen-task evaluation:} we further add 2 datasets for crowd counting to evaluate the generalization ability to unseen tasks. More detailed distributions of these evaluation datasets are presented in Tab.~\ref{tab:downstream}.


\noindent \textbf{Evaluation Protocols.} For each evaluation scenario, we expect a good representation can generalize to specific human-centric tasks without updating the feature extractor or being a good starting point when adapted to any specific human-centric tasks by finetuning. Therefore, we present three evaluation protocols for the experiments presented in Sec.~\ref{sec:exp}.

\noindent \underline{\emph{Full Finetuning.}} Full Finetuning evaluates the generalization ability when pretraining models serve as training starting points. In this case, we load the pretrained backbone and finetune all layers by supervision from downstream tasks.

\noindent \underline{\emph{Head Finetuning.}} Head Finetuning is very similar to linear probing~\cite{he2020momentum} in self-supervised learning on natural image classification. It evaluates the generalization ability of pretrained models without updating. Therefore, we keep the pretrained backbone frozen and learn simple task heads for downstream datasets.

\noindent \underline{\emph{Partial Finetuning.}} Partial Fintuning is a setting between head finetuning and full finetuning~\cite{he2022masked}, which finetunes the last several layers while freezing the others. This evaluation protocol can take advantage of both full finetuning and head finetuning, \emph{i.e.,} it can efficiently evaluate the opportunity of pursuing strong but non-linear features.

\begin{figure*}
    \centering
    \includegraphics[width=0.9\linewidth]{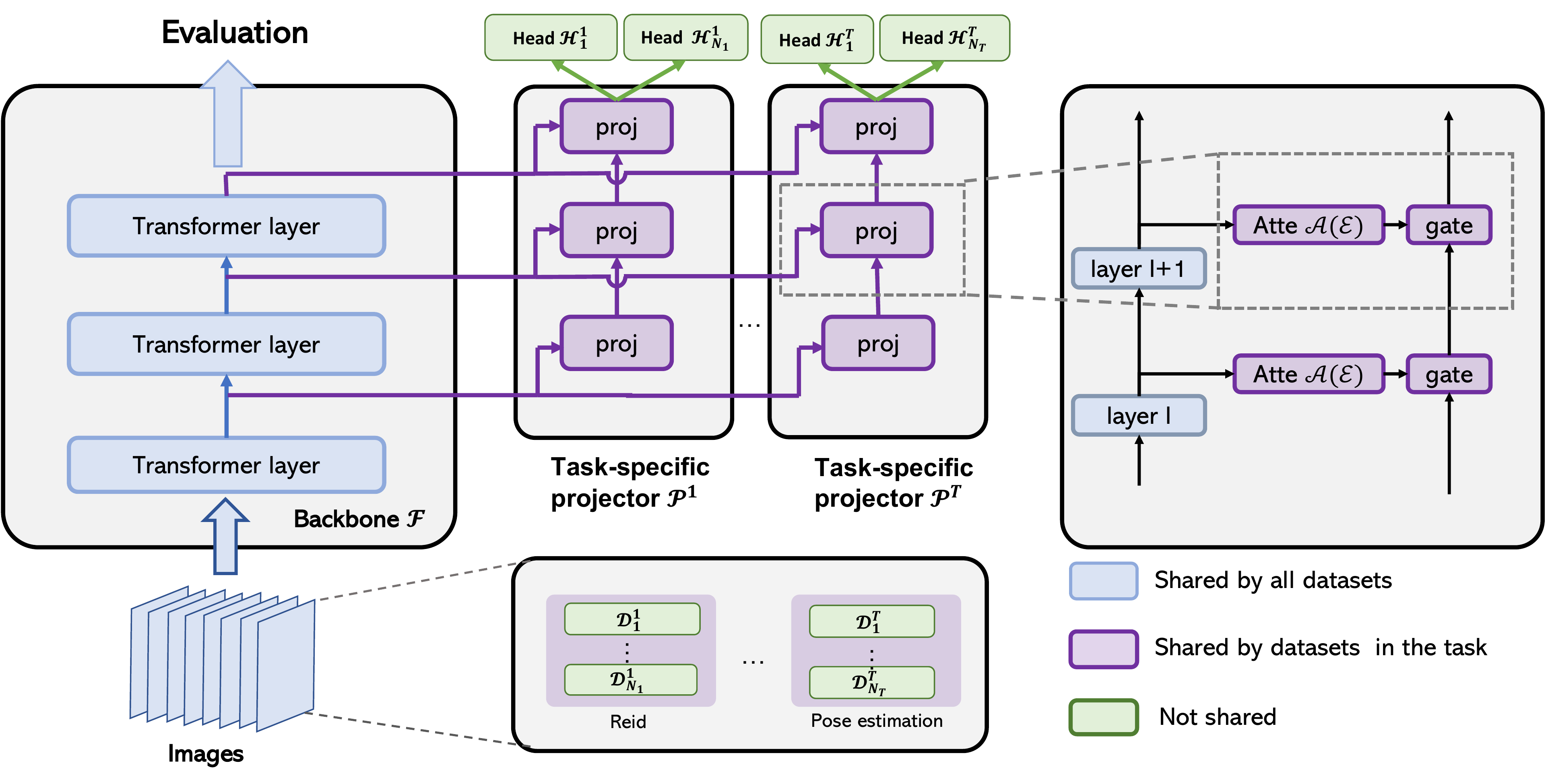}
    \caption{Overview of our proposed pretraining method, PATH. Images from various datasets are fed into the same backbone to extract the general features, and then the task-specific projector attends to the task-specific features from the general features. The dataset-specific head is imposed to predict dataset-specific results, which are fed into the loss function for training. In this framework, we adopt a hierarchical weight-sharing strategy, where parameters in the backbone are shared among all datasets, parameters in a task-specific projector is shared among all datasets belonging to the same task, and the parameters for the heads are not shared.}
    \label{fig:framework}
    \vspace{-1.5em}
\end{figure*}

\section{Methodology}
We now introduce our proposed projector-assisted pretraining method (PATH) with hierarchical weight sharing. Our method is motivated by~\cite{wang2022revisiting}, which reveals that inserting an MLP projector before the objective function can significantly increase the generalization ability of supervised pretraining. To avoid task conflicts among various tasks, we improve this method by inserting task-specific projectors between the backbone and the head of every dataset and designing a new hierarchical weight-sharing strategy. Concretely, the projectors are very lightweight modules composed of an attention module and a gating module (Sec.~\ref{sec:projector_design}). The hierarchical weight-sharing strategy enforces that parameters of backbone, projectors, and heads are shared among all datasets across different tasks, shared among datasets in the same task, and not shared, respectively (Sec.~\ref{sec:hierarchical_weight_sharing}). As such, we expect the backbone to learn general representations of all human-centric tasks, the projector to attend to the task-specific features from the general representations, and the head to supervise the network optimization by the annotations of every dataset. To evaluate the generalization ability of the pretrained backbone, we discard the projectors and heads, using the backbone only.

\subsection{Hierarchical Weight Sharing} \label{sec:hierarchical_weight_sharing}
We design a hierarchical weight-sharing strategy to reduce task conflicts among various annotations. Specifically, our model consists of three components: a single backbone shared by all datasets, $T$ task-specific projectors shared by all datasets in the same task, and $N$ dataset-specific heads that are not shared, where $N=N_1\!+\!N_2\!+\!...\!+\!N_T$ is the number of datasets in the pretraining dataset and $N_t$ is the number of datasets in the $t$-th task. 

\noindent \textbf{Backbone.}
The backbone $\mathcal{F}$ is implemented by a plain vision transformer~\cite{dosovitskiy2020image} in the experiments. The parameters of the backbone are shared by all datasets regardless of tasks. 

\noindent \textbf{Task-specific projector.}
Each task-specific projector $\mathcal{P}^t$ consists of sets of attention modules and gating modules, which link with the backbone $\mathcal{F}$, where $t \le T$ and $T$ is the number of tasks. Since the parameters of the task-specific projector are shared among the datasets with the same task, the attention modules in the task-specific network can be considered as selecting features from the shared backbone network, whilst the shared backbone network learns a compact global feature pool across all datasets. 

\noindent \textbf{Dataset-specific head.} To tackle the possible data distribution shift in different datasets, we still preserve the dataset-specific head $\mathcal{H}^t_j$, whose parameters are not shared. Here, $\mathcal{H}^t_j$ is the $j$-th dataset in the $t$-th task.


Figure~\ref{fig:framework} shows a detailed visualization of our PATH. The detailed pipeline is described as follows.

\emph{Step1: Extract the general features of images in the pretraining dataset.} Given an image $\mathbf{x}$ sampled from $\mathcal{D}^t_{j}$ which is the $j$-th dataset in the $t$-th task in the pretraining dataset, extract the intermediate and final feature maps $\mathbf{F}$ by the backbone, which will be fed into the projectors. 

\emph{Step2:} \emph{Attend the task-specific features by the task-specific projector (Sec.~\ref{sec:projector_design}).} Given the feature maps $\mathbf{F}$ from the backbone, we attend the task-specific features $\mathbf{p}=\mathcal{P}^t(\mathbf{F})$ by the $t$-th task-specific projector.

\emph{Step3:} \emph{Calculate the activations by dataset-specific heads, losses by the activations, and optimize the parameters of the backbone, the projector, and the head simultaneously by backward propagation (Sec.~\ref{sec:loss}).}

During the evaluation stage, we discard the projectors and evaluate the generalization ability of the backbone $\mathcal{F}$ using the protocols in Sec.~\ref{sec:evaluation_protocol}.


\subsection{Design of Task-specific Projector} \label{sec:projector_design}
The task-specific projector is designed to attend to task-specific features from backbone outputs, by applying an alternating chain of the attention module and gating module to the features in the shared backbone. Given an image $\mathbf{x}$ sampled from the $j$-th dataset in the $t$-th task, \emph{i.e.,} $\mathcal{D}^t_j$ and its intermediate feature maps $\mathbf{f}_l$ in the $l$-th transformer block, we leverage a squeeze-and-excitation layer~\cite{hu2018squeeze} to implement the channel attention and a self-attention module~\cite{vaswani2017attention} to implement spatial attention to generate the attended feature maps $\mathbf{z}_l$. Mathematically, $\mathbf{z}_l = \mathcal{A}^t(\mathcal{E}^t(\mathbf{f}_l)),$
where $\mathcal{A}^t$ and $\mathcal{E}^t$ respectively denote standard self-attention blocks~\cite{vaswani2017attention} and squeeze-and-excitation blocks~\cite{hu2018squeeze} for the $t$-th task. We will detail the structure of the Squeeze-and-excitation module and self-attention module in the supplementary materials.

In order to effectively aggregate features from different layers of the backbone, we design a gating module to dynamically aggregate features from different layers. Specifically, given the feature map $\mathbf{z}_l$ after the attention module and the gated feature maps $\mathbf{p}_{l-1}$ in the $(l-1)$-th layer, the gating function aggregates features as follows:
\begin{equation} \label{eq:gating}
    \mathbf{p}_l = \mu_l \mathbf{z}_l + (1-\mu_l) \mathbf{p}_{l-1},
\end{equation}
where $\mathbf{p}_1=\mathbf{z}_1$,  $\mu_l = \sigma(\alpha_l/T)$ is a gate parameterized with a learnable zero-initilalized scalar $\alpha_i$ and temperature $T$(=0.1), and $\sigma$ is the sigmoid function. By iteratively computing Eq.~\ref{eq:gating} from $l=1$ to $L$, we generate the final feature maps \emph{i.e.,} $\mathbf{p}=\mathbf{p}_L$ for the dataset head.

\subsection{Dataset-specific Head and Objective Functions} \label{sec:loss}
Dataset-specific heads aim at transforming task-specific features into activations for computing losses of every dataset. In general multi-dataset learning with $N$ datasets, the features $\mathbf{P}_i$ after the projector of all images $\mathbf{X}_i$ and labels $\mathbf{Y}_i$, $i=1, 2,..., N$ in $i$-th dataset, the objective function is defined as 
$\mathcal{L} = \sum_{i=1}^{N}\lambda_i \mathcal{L}_i(\mathbf{Z}_i, \mathbf{Y}_i),$
where $\mathbf{Z}_i$ is the activation generated by the dataset-specific head.
This is the linear combination of dataset-specific losses $\mathcal{L}_i$ with task weightings $\lambda_i$. In this paper, we follow some basic head and loss function designs of all pretraining tasks we include. Specifically, we follow the head and loss function designs in VitPose~\cite{xu2022vitpose} for pose estimation, in TransReID~\cite{he2021transreid} for person ReID, in Segformer~\cite{xie2021segformer} for human parsing, in Anchor Detr~\cite{wang2022anchor} for pedestrian detection, in Label2Label~\cite{li2022label2label} for pedestrian attribute recognition, and in DR.VIC~\cite{han2022dr} for crowd counting. More details of these head and loss designs will be elaborated in the supplementary materials.

\subsection{Technical Details}
\noindent \textbf{Replacing all Batchnorm with Layernorm in pose and parsing decoders.} Generally, the original feature normalization method in pose estimation and human parsing tasks is batch normalization with CNN backbone, which renders the model to learn powerful feature distribution from the statistics of batch inputs when trained on a single domain. However, in HumanBench, each task has different datasets, which may have domain gaps, resulting in inaccurate normalization statistics when the dataset-specific head is fed with features from the task-share projector. To reduce the inaccurate statistics, we replace the Normalization method from BatchNorm\cite{ioffe2015batch} to LayerNorm\cite{ba2016layer} and experimentally find that it can improve feature representation. 

\noindent \textbf{Sharing Positional Embedding among All Datasets.} In HumanBench, the input image size of different tasks varies largely, resulting in different numbers of patch embeddings and positional embeddings after projecting an image to patch embedding. As a result, different tasks can not share positional embeddings when the model is trained in a distributed manner. To tackle this problem, we parameterize  positional embeddings as 224$\times$224 in all tasks and interpolate its size according to the actual input image size of each dataset during the pretraining stage.

\section{Experiment} \label{sec:exp}

\subsection{Experimental Setup}
The backbone used for experiments is the plain ViT-base. It has 12 transformer blocks with the dimension of patch embedding 768 and 12 attention heads. In the pre-train stage, each GPU is responsible for one dataset independently for training in a distributed manner. We use Adafactor\cite{shazeer2018adafactor} optimizer with base learning rate of 5e-4 and weight decay of 0.05. We linearly warmup the learning rate from 1e-7 to 5e-4 for the first 1500 iteration steps. Step learning rate decay of 0.5 is used in 50\%, 75\%, 95\% iterations. For the ViT-Base encoder, we set a layer-wise learning rate decay of 0.75 for 12 transformer blocks and the model is trained for 80k iterations.

\subsection{Experimental Results}

\input{table_main}

As detailed in Sec.~\ref{sec:evaluation_protocol}, we implement in-dataset evaluation, out-of-dataset evaluation, and unseen-task evaluation on \textbf{HumanBench}. Both in-dataset evaluation, out-of-dataset evaluation include 5 human-centric tasks, \emph{i.e.,} person ReID, pose estimation, human parsing, pedestrian attribute recognition, and pedestrian detection. 
The unseen downstream task which is not in the pretraining tasks, \emph{i.e,} crowd counting, evaluates the generalization ability to unseen tasks. The compared methods are the state-of-the-art methods of each task and two popular pretraining models, \emph{i.e.,} MAE \cite{he2022masked} and CLIP \cite{radford2021learning}. 
MAE is a newly proposed vision self-supervised pretraining method. Pretrained on ImageNet-1K, MAE achieves excellent results for many visual tasks. CLIP learns generic and transferable representations from a dataset of 400 million (image, text) pairs. 
We summarize our experimental results with 3 evaluation scenarios and 3 evaluation protocols in Tab.~\ref{tab_main_res}.

\noindent \textbf{In-dataset Evaluation.} \label{sec:seen}
In-dataset evaluation quantifies the ability of the pretraining method when it is evaluated on the data with similar data distribution and pretrained tasks. 
As shown in Tab.~\ref{tab_main_res}, compared with SoTA methods used in different papers for their specific tasks, our HumanBench with full finetuning achieves better performance on 8 datasets. Specifically, for human parsing, we improve the current state-of-the-art results by \textbf{+2.5\%} mIOU, \textbf{+1.1\%} mIOU and \textbf{+1.2\%} mIOU on Human3.6M, LIP and CIHP, respectively. We also improve the person ReID by \textbf{+4.9\%} mAP on CUHK03 datasets. We notice our results are lower than PASS~\cite{zhu2022pass} on Market1501 and MSMT, probably because PASS uses techniques, \emph{i.e.,} part models~\cite{wang2018learning,sun2018beyond}, that are time-consuming (120 hours using 8 A100 GPUs) but specifically effective for ReID. Besides, we improve pose estimation by \textbf{+3.0\%} AP, \textbf{-1.2\%} MR$^{-2}$$(\downarrow)$ on AIC and Human3.6m, respectively. Furthermore, we improve pedestrian attribute recognition \textbf{+1.5\%} mA and \textbf{+0.2\%} mA on PA-100K and RAPv2 datasets, respectively.

To evaluate the generalization of different methods when all backbone parameters or most of the backbone parameters are frozen, we further evaluate our HumanBench with head finetuning and partial finetuning with 100\% of the downstream data. We observe that our method with only head finetuning can be on par with and even surpasses the SoTAs in 12 seen datasets, such as \textbf{-1.2\%} heavy occluded MR$^{-2}$$(\downarrow)$ \textbf{+1.6\%} mIOU on Human3.6m pose estimation and human parsing tasks. Our HumanBench with partial finetuning performs better than full finetuning in 2 Pedestrian Attribute Recognition datasets (PA-100K and RapV2) of 12 seen datasets, possibly because these two datasets have fewer data. 

We also use ViT-Large to verify the model scalability of our method PATH on HumanBench in Tab.~\ref{tab_main_res}. Results show that the results with a large backbone under partial finetuning can further achieve considerable gains over the best ViT-Base results, showing the promising scalability of our proposed pretraining method PATH on HumanBench.

\noindent \textbf{Out-of-dataset Evaluation.}
To quantify the generalization ability of  pretrained models on tasks with potentially different data distribution but the same task in the pretraining dataset, we implement out-of-dataset evaluations on 5 datasets, \emph{i.e.,} ATR, SenseReID, Caltech, MPII, PETA, one dataset for each pretraining task. As shown in Tab.~\ref{tab_main_res}, our pretraining method PATH performs better than previous methods in 4 of 5 unseen datasets and comparable in the remaining one. To be concrete, our method improves by \textbf{+0.1\%}pACC, \textbf{+4.4\%}Top1 accuracy, \textbf{-0.5\%} heavy occluded MR$^{-2}$$(\downarrow)$ and \textbf{+2.7\%} mA on ATR (human parsing), SenseReID (person ReID), Caltech (pedestrian detection) and PETA datasets (pedestrian attribute recognition), respectively.
These significant and consistent performance gains across different datasets verify the generalization ability of our pretrained model to tasks with potentially different data distributions. We also observe the results when we only finetune the last two layers are already on par or even better than the results by full finetuning. Especially, the results of SenseReID, Caltech and PETA by partial finetuing are better than that of full finetuning by \textbf{+0.5\%}Top1 accuracy, \textbf{-1.8\%}heavy occluded MR$^{-2}$$(\downarrow)$ and \textbf{+1.8\%}mA, showing the good generalization of our pretrained models and its easy deployment in the real world. Similar to the results in the \emph{out-of-dataset evaluation}, partial finetuing performs better than full finetuning when the dataset is small in Caltech (4250 images) and PETA (9500 images). Therefore, partial finetuning can be a choice when the downstream dataset has few samples.


\noindent \textbf{Unseen-task Evaluations.} \label{sec:unseen}
To evaluate the generalization ability to unseen tasks, we construct an \emph{unseen task evaluation}, in which the task is not involved in the pretraining tasks, \emph{i.e,} crowd counting. As presented in Tab.~\ref{tab_main_res}, our pretrained model achieves significant performance gains than the MAE pretrained model by \textbf{-10.4\%} MSE$(\downarrow)$ and \textbf{-4.7\%} MSE$(\downarrow)$. Our HumanBench improves previous SoTAs that are specially designed for crowd counting by \textbf{-2.6\%} MSE$(\downarrow)$ and \textbf{-0.2\%} MSE$(\downarrow)$ on ShTech PartA and PartB datasets, respectively. These consistent improvements validate the generalization ability of our learned representations.

\noindent \textbf{Comparsion with MAE and CLIP Models.} We also compare our pretrained method with other popular pretrained models, \emph{i.e.,} MAE and CLIP, on our proposed HumanBench. In Tab.~\ref{tab_main_res}, we find our pretraining method performs considerably better than CLIP and MAE under the full finetuning evaluation protocol on all tasks. Interestingly, the performance of CLIP\footnote{We carefully tune the learning rate, drop path rate, and weight decay of CLIP pretrained ViT-B and report the best results we have ever achieved.} is lower than MAE, which shows that more data on natural images and languages may not naturally benefit a variety of human-centric tasks, which empirically validates the importance of our HumanBench for further research on human-centric pretraining. 


\input{table_ablation}

\subsection{Ablation Study} \label{sec:ablation}
 Due to the significant computation cost with the large-scale full datasets, as summarized in Table 1 in \textbf{Supplementary Material}, we sample a subset containing a similar number of images as ImageNet-1K ($\sim$ 1.28 M) from the full training set. We pretrain our models on this subset to verify the effectiveness of our designs by default in this section, and implement 4 in-dataset evaluations (PA-100K, LIP, Market1501, MSMT) and 3 out-of-dataset evaluations (Caltech, PETA, MPII) full dataset finetuning.

\noindent \textbf{Effectiveness of hierarchical weight sharing.}
To verify the effectiveness of our hierarchical weight sharing, we adapt the three projector share strategies: (1) all shared projector (\textbf{A}): sharing the projector parameters across all the tasks and datasets (Table \ref{tab_ablation} (a) ); (2) task-shared projector (\textbf{T}): sharing the projector parameters across all datasets in a single task, while maintaining an independent projector for each task (Table \ref{tab_ablation} (c) ); and (3) specific projector (\textbf{S}): maintaining an independent projector for each dataset (Table \ref{tab_ablation} (b) ).  The results show that the task-shared projector is better than the other two. We speculate that the projector is the core component to map the general human-centric features to task-specific features. Therefore, all datasets in the same task are supposed to share the same mapping functionality while different tasks 
should operate differently due to the existing task gaps.


\noindent \textbf{Effectiveness of shared positional embedding.}
Experiments (c) and (d) in Tab.~\ref{tab_ablation} ablate whether positional embeddings are shared or not across the different tasks. The results show that shared positional embedding helps to learn general human-centric representations and leads to \textbf{+0.3\%} improvement on average when the five tasks are considered. Since the backbone is shared, we conjecture that independent positional embedding would cause inconsistency between different tasks, create barriers in learning shared backbone across tasks, and result in difficulties in learning the model.


\input{table_ablation_data}

\noindent \textbf{Comparison with self-supervised pretraining methods.} 
As shown in Table \ref{tab_ablation_data}, we first ablate the effectiveness of our dataset on downstream tasks. With almost the same number of images, the MAE pretrained on our subset for 800 epochs surpasses ImageNet pretrained MAE by +4.4\%, which shows that by combining the diverse human-centric data across various human-centric tasks, our dataset is more suitable to learn human-centric features. Second, pretrained on our subset, our supervised pretraining method, i.e. PATH, performs better than both MAE (800 epochs) and MOCOv3 (800 epochs) by +1.8\%. Different from MAE and MOCOv3 which ignore the general properties of the human body and the potential association between the data in different tasks, our PATH is designed to capture the potential complementary knowledge between different tasks, leading to learning more general human-centric representations to improve the performance on various human-centric tasks.

\section{Conclusion}
In this paper, we investigate the opportunities and challenges in pretraining on various human-centric tasks, and propose a new HumanBench with the existing publicly available datasets. Based on HumanBench, we design a projector-assisted pretraining with hierarchical weight sharing (PATH) to learn human-centric information from annotations with different granularities. We hope our HumanBench can facilitate future works such as unified network structure design and multi-task/supervised/self-supervised learning methods on a broad variety of human-centric tasks.

{\small
\bibliographystyle{ieee_fullname}
\bibliography{PaperForReview}
}

\clearpage
\appendix

\section{Details of HumanBench}
In the main text, we briefly introduce the number of images and number of tasks in the pretraining dataset of HumanBench. For the evaluation of HumanBench, we introduce the evaluation scenario and evaluation protocols. In this section, we present detailed information on the pretraining dataset and evaluation dataset and discuss the ethical issues of these datasets.
\subsection{Dataset Statistics of HumanBench}
HumanBench collects 37 publicly available datasets of 5 human-centric tasks, including person ReID, human parsing, pose estimation, pedestrian detection, and pedestrian attribute. More details can be seen in Table~\ref{tab:dataset statistics}.  The existing distribution of datasets includes large numbers of human-centric cropped images in ReID, video frames in person pose estimation, and human parsing. In particular, to avoid information reduction, we select a single frame from every 8 video frames. Particularly, except for using training images in all datasets, we also use all/partial test images in some datasets. Specifically, for the person Reid task, we use all test images in LaST and partial test images in the PRCC dataset; for the human parsing task, we only use train images and publicly released images in DeepFashion($\sim$half of the dataset reported in~\cite{liu2016deepfashion}). For the pedestrian detection dataset, we remove the images in which there is no person. For the pose estimation datasets, we only use train images. For the pedestrian attribute recognition dataset, we only use partial test images in the UAV-Human dataset and do not contain test images in other pedestrian attribute recognition datasets. All the images in the pretraining dataset have been de-duplicated with the testing datasets to be a meaningful benchmark of our HumanBench.

\subsection{Discussion of Ethical Issues}
The usage of HumanBench might bring several risks, such as privacy, and problematic content. We discuss these risks and their mitigation strategies as follows. 

\paragraph{Copyright.} All images in this paper and dataset are collected by publicly available. We claim the dataset:
\begin{itemize}
    \item Copy and redistribute the material in any medium or format.
    \item Remix, transform and build on the material for any purpose, even commercially.
\end{itemize}
Referring to OmniBenchmark~\cite{zhang2022benchmarking}, MS-COCO~\cite{lin2014microsoft}, Kinetics-700~\cite{carreira2019short}, we only present the lists of URLs and their corresponding meta information to our HumanBench. 

\subsection{Details of HumanBench-Subset}
Due to the significant computational cost when we pretrain the model on the full dataset, we select 17 subsets from 37 full datasets for ablation study, which contains 1,270,186 images as a similar number with ImageNet-1K($\sim$1.28M). Table~\ref{tab:dataset statistics} summarizes the statistics of HumanBench-Subset. For the person ReID task, we select widely-used Market1501 and CUHK03 datasets, and the clothes-changing ReID dataset PRCC, forming a total of 38,197 images. For the human parsing task, we select widely used Human3.6M, LIP, CIHP, VIP datasets and one clothes parsing dataset, \emph{i.e.,} ModaNet, with a total of 192,124 images. For the pose estimation task, we select widely-used COCO, AIC, and PoseTrack datasets with a total of 748,812 images. For the attribute task, we select PA-100K, RAPv2, and Market1501-Attribute datasets with a total of 170,879. Due to the significant resource cost, for the pedestrian detection task, we only select one widely used dataset CrowdHuman.

\begin{table*}[t]
  \centering
  \footnotesize
  \caption{Dataset statistics of pretraining datasets}
    \begin{tabular}{c|c|cc|c|cc}
    \toprule
    Partition & Task  & Name  & Number of images / samples & Task  & Name  & Number of images / samples \\
    \hline
    \multirow{19}[12]{*}{Full} & \multirow{7}[4]{*}{ReID} & Market1501~\cite{zheng2015scalable} & 12,936  & Detection & WIDER Pedestrian~\cite{loy2019wider} & 57,999  \\
\cline{5-7}          &       & CUHK03~\cite{li2014deepreid} & 7,365  & \multirow{11}[4]{*}{Pose} & COCO~\cite{lin2014microsoft}  & 262,465  \\
          &       & MSMT~\cite{wei2018person} & 30,248  &       & AIC~\cite{wu2019large}   & 378,374  \\
          &       & LaST~\cite{shu2021large}  & 71,248  &       & PoseTrack~\cite{andriluka2018posetrack} & 107,973  \\
          &       & PRCC~\cite{yang2019person}  & 17,896  &       & JRDB~\cite{vendrow2022jrdb}  & 310,035  \\
          &       & DGMarket~\cite{zheng2019joint} & 128,306  &       & MHP~\cite{li2017multiple}   & 41,128  \\
          &       & LUPerson-NL~\cite{fu2021unsupervised} & 5,178,420  &       & UppenAction~\cite{zhang2013actemes} & 163,839  \\
\cline{2-4}          & \multirow{7}[4]{*}{Parsing} & Human3.6M~\cite{h36m_pami} & 62,668  &       & Halpe~\cite{fang2022alphapose} & 41,712  \\
          &       & LIP~\cite{gong2017look}   & 30,462  &       & 3dpw~\cite{von2018recovering}  & 74,620  \\
          &       & CIHP~\cite{gong2018instance}  & 28,280  &       & MPI-INF-3DHP~\cite{mehta2017monocular} & 1,031,701  \\
          &       & VIP~\cite{zhou2018adaptive}   & 18,469  &       & Human3.6M~\cite{h36m_pami} & 312,187  \\
          &       & Paper Doll~\cite{yamaguchi2013paper} & 1,035,825  &       & AIST++~\cite{li2021ai} & 1,015,257  \\
\cline{5-7}          &       & DeepFashion~\cite{liu2016deepfashion} & 191,961  & \multirow{6}[4]{*}{Attribute} & PA100K~\cite{liu2017hydraplus} & 90,000  \\
          &       & ModaNet~\cite{zheng2018modanet} & 52,245  &       & RAPv2~\cite{li2018richly} & 67,943  \\
\cline{2-4}          & \multirow{5}[4]{*}{Detection} & CrowdHuman~\cite{shao2018crowdhuman} & 15,000  &       & HARDHC~\cite{li2016human} & 28,336  \\
          &       & WiderPerson~\cite{zhang2019widerperson} & 9,000  &       & UVA-Human~\cite{li2021uav} & 16,183  \\
          &       & COCO-person~\cite{lin2014microsoft} & 64,115  &       & Parse27k~\cite{PARSE27k} & 27,482  \\
          &       & EuroCity Persons~\cite{braun2018eurocity} & 21,795  &       & Market1501-Attribute~\cite{zheng2015scalable} & 12,936  \\
\cline{5-7}          &       & CityPersons~\cite{zhang2017citypersons} & 2,778  & \textbf{Total } & \multicolumn{2}{c}{\textbf{11,019,187}} \\
    \hline\hline
    \multirow{8}[8]{*}{Subset} & \multirow{3}[2]{*}{ReID} & Market1501~\cite{zheng2015scalable} & 12,936  & \multirow{3}[2]{*}{Pose} & COCO~\cite{lin2014microsoft}  & 262,465  \\
          &       & CUHK03~\cite{li2014deepreid} & 7,365  &       & AIC~\cite{wu2019large}   & 378,374  \\
          &       & PRCC~\cite{yang2019person}  & 17,896  &       & PoseTrack~\cite{andriluka2018posetrack} & 107,973  \\
\cline{2-7}          & \multirow{5}[6]{*}{Parsing} & Human3.6M & 62,668  & Detection & CrowdHuman~\cite{shao2018crowdhuman} & 15,000  \\
\cline{5-7}          &       & LIP~\cite{gong2017look}   & 30,462  & \multirow{3}[2]{*}{Attribute} & PA100K~\cite{liu2017hydraplus} & 90,000  \\
          &       & CIHP~\cite{gong2018instance}  & 28,280  &       & RAPv2~\cite{li2018richly} & 67,943  \\
          &       & VIP~\cite{zhou2018adaptive}   & 18,469  &       & Market1501-Attribute~\cite{zheng2015scalable} & 12,936  \\
\cline{5-7}          &       & ModaNet~\cite{zheng2018modanet} & 52,245  & \textbf{Total } & \multicolumn{2}{c}{\textbf{1,165,012 }} \\
    \hline
    \end{tabular}%
  \label{tab:dataset statistics}%
\end{table*}%

\begin{table*}[t]
  \scriptsize
  \centering
  \caption{Results of the publicly released MAE and CLIP on HumanBench.}
    \begin{tabular}{cl|cccccccccc}
    \toprule
          &       & \multicolumn{4}{c}{Human Parsing} & \multicolumn{4}{c}{Person ReID} & \multicolumn{2}{c}{Pedestrian  Detection} \\
\cmidrule{3-12}          &       & H3.6M & LIP   & CIHP  & \cellcolor[rgb]{ .949,  .949,  .949} ATR & Market1501 & MSMT  & CUHK03 & \cellcolor[rgb]{ .949,  .949,  .949} SenseReID & CrowdHuman & \cellcolor[rgb]{ .949,  .949,  .949} Caltech ($\downarrow$) \\
    \midrule
    \multirow{9}[6]{*}{ViT-B} & MAE   & 62.0  & 57.2  & 61.7  & \cellcolor[rgb]{ .949,  .949,  .949} 97.4  & 79.2  & 51.5  & 65.8  & \cellcolor[rgb]{ .949,  .949,  .949} 44.6  & 89.6  & \cellcolor[rgb]{ .949,  .949,  .949} 48.1  \\
          & MAE (Head FT) & 40.4  & 31.7  & 37.1  & \cellcolor[rgb]{ .949,  .949,  .949} 94.4  &    &    &    &    & 75.7  & \cellcolor[rgb]{ .949,  .949,  .949} 66.2  \\
          & MAE (Partial FT) & 50.5  & 42.0  & 48.0  & \cellcolor[rgb]{ .949,  .949,  .949} 96.4  & 43.8  & 22.5  & 33.2  & \cellcolor[rgb]{ .949,  .949,  .949} 21.2  & 82.6  & \cellcolor[rgb]{ .949,  .949,  .949} 70.2  \\
\cmidrule{2-12}          & CLIP  & 58.2  & 53.4  & 61.7  & \cellcolor[rgb]{ .949,  .949,  .949} 97.0  & 78.6  & 53.6  & 66.9  & \cellcolor[rgb]{ .949,  .949,  .949} 43.6  & 82.1  & \cellcolor[rgb]{ .949,  .949,  .949} 78.6  \\
          & CLIP (Head FT) & 28.4  & 11.7  & 14.2  & \cellcolor[rgb]{ .949,  .949,  .949} 85.8  &    &    &    &    & 33.2  & \cellcolor[rgb]{ .949,  .949,  .949} 98.5  \\
          & CLIP (Partial FT) & 32.1  & 24.8  & 30.0  & \cellcolor[rgb]{ .949,  .949,  .949} 90.8  & 34.5  & 10.9  & 15.2  & \cellcolor[rgb]{ .949,  .949,  .949}25.3  & 28.4  & \cellcolor[rgb]{ .949,  .949,  .949} 97.1  \\
\cmidrule{2-12}          & Ours (FT) & 65.0  & 61.4  & 66.8  & \cellcolor[rgb]{ .949,  .949,  .949} 97.5  & 89.5  & 69.1  & 82.6  & \cellcolor[rgb]{ .949,  .949,  .949} 56.8  & 90.6  & \cellcolor[rgb]{ .949,  .949,  .949} 30.1  \\
          & Ours (Head FT) & 64.1  & 59.9  & 63.3  & \cellcolor[rgb]{ .949,  .949,  .949} 97.1  &    &    &    & \cellcolor[rgb]{ .949,  .949,  .949}  & 90.0  & \cellcolor[rgb]{ .949,  .949,  .949} 31.1  \\
          & Ours (Partial FT) & 63.7  & 60.0  & 63.1  & \cellcolor[rgb]{ .949,  .949,  .949} 97.2  & 88.7  & 66.1  & 79.5  & \cellcolor[rgb]{ .949,  .949,  .949} 57.2  & 90.9  & \cellcolor[rgb]{ .949,  .949,  .949} 28.3  \\
\cmidrule{1-12}    \multirow{3}[4]{*}{ViT-L} & MAE (Partial FT) & 21.94 & 10.74 & 11.98 & \cellcolor[rgb]{ .949,  .949,  .949} 85.8  & 44.9  & 23.6  & 64.5  & \cellcolor[rgb]{ .949,  .949,  .949} 23.5  & 10.1  & \cellcolor[rgb]{ .949,  .949,  .949} 99.8  \\
          & CLIP (Partial FT) & 14.68 & 10.05 & 4.37  & \cellcolor[rgb]{ .949,  .949,  .949} 81.4  & 33.6  & 12.9  & 16.8  & \cellcolor[rgb]{ .949,  .949,  .949} 27.6  & 6.5   & \cellcolor[rgb]{ .949,  .949,  .949} 99.4  \\
\cmidrule{2-12}          & Ours (Partial FT) & 66.2  & 62.6  & 67.5  & \cellcolor[rgb]{ .949,  .949,  .949} 97.4  & 91.8  & 74.7  & 86.0  & \cellcolor[rgb]{ .949,  .949,  .949} 66.8  & 90.8  & \cellcolor[rgb]{ .949,  .949,  .949} 28.7  \\
    \midrule
          &       & \multicolumn{4}{c}{Pose Eestimation} & \multicolumn{3}{c}{Pedestrian Attribute Recognition} & \multicolumn{2}{c}{Counting (unseen task)} &  \\
\cmidrule{3-11}          &       & COCO  & H3.6M ($\downarrow$) & AIC   & \cellcolor[rgb]{ .949,  .949,  .949} MPII & PA100K & Rapv2 & \cellcolor[rgb]{ .949,  .949,  .949} PETA & ShTech PartA ($\downarrow$) & ShTech PartB ($\downarrow$) &  \\
    \multirow{9}[4]{*}{ViT-B} & MAE   & 75.8  & 8.2   & 31.8  & \cellcolor[rgb]{ .949,  .949,  .949} 90.1  & 82.3  & 80.8  & \cellcolor[rgb]{ .949,  .949,  .949} 84.6  & 102.1  & 15.5  &  \\
          & MAE (Head FT) & 60.9  & 7.9   & 19.2  & \cellcolor[rgb]{ .949,  .949,  .949} 84.6  & 55.9  & 55.1  & \cellcolor[rgb]{ .949,  .949,  .949} 61.4  & 156.2  & 32.5  &  \\
          & MAE (Partial FT) & 69.2  & 8.0   & 26.9  & \cellcolor[rgb]{ .949,  .949,  .949} 88.5  & 78.5  & 82.3  & \cellcolor[rgb]{ .949,  .949,  .949} 85.0  & 135.6  & 26.8  &  \\
\cmidrule{2-12}          & CLIP  & 74.4  & 9.9   & 31.1  & \cellcolor[rgb]{ .949,  .949,  .949} 88.1  & 76.1  & 77.0  & \cellcolor[rgb]{ .949,  .949,  .949} 81.2  & 117.9  & 16.3  &  \\
          & CLIP (Head FT) & 28.4  & 11.7  & 14.2  & \cellcolor[rgb]{ .949,  .949,  .949} 85.8  & 51.3  & 51.5  & \cellcolor[rgb]{ .949,  .949,  .949} 54.9  & 198.5  & 36.8  &  \\
          & CLIP (Partial FT) & 32.1  & 24.8  & 30.0  & \cellcolor[rgb]{ .949,  .949,  .949} 90.8  & 72.8  & 76.3  & \cellcolor[rgb]{ .949,  .949,  .949} 81.5  & 168.6  & 32.3  &  \\
\cmidrule{2-12}          & Ours (FT) & 76.3  & 6.2   & 35.0  & \cellcolor[rgb]{ .949,  .949,  .949} 93.3  & 85.0  & 81.2  & \cellcolor[rgb]{ .949,  .949,  .949} 88.0  & 91.7  & 10.8  &  \\
          & Ours (Head FT) & 75.2  & 6.1   & 31.6  & \cellcolor[rgb]{ .949,  .949,  .949} 92.7  & 77.4  & 72.4  & \cellcolor[rgb]{ .949,  .949,  .949} 79.0  & 97.6   &  13.8  &  \\
          & Ours (Partial FT) & 76.0  & 6.1   & 33.3  & \cellcolor[rgb]{ .949,  .949,  .949} 93.0  & 86.9  & 83.1  & \cellcolor[rgb]{ .949,  .949,  .949} 89.8  & 94.3   & 14.0  &  \\
\cmidrule{1-12}
    \multirow{3}[3]{*}{ViT-L} & MAE (Partial FT) & 76.5  & 6.6   & 34.2  & \cellcolor[rgb]{ .949,  .949,  .949} 92.2  & 67.2  & 51.3  & \cellcolor[rgb]{ .949,  .949,  .949} 51.7  & 141.3  & 27.9  &  \\
          & CLIP (Partial FT) & 62.3  & 10.8  & 14.8  & \cellcolor[rgb]{ .949,  .949,  .949} 73.0  & 55.0  & 58.5  & \cellcolor[rgb]{ .949,  .949,  .949} 50.1  & 167.6  & 30.6  &  \\
\cmidrule{2-12}          & Ours (Partial FT) & 77.1  & 5.8   & 36.3  & \cellcolor[rgb]{ .949,  .949,  .949} 93.7  & 90.8  & 87.4  & \cellcolor[rgb]{ .949,  .949,  .949} 90.7  & 91.3  & 11.5  &  \\
    \bottomrule
    \end{tabular}%
  \label{tab:more_results}%
\end{table*}%

\section{More Results of MAE and CLIP on HumanBench}
In this section, we provide more results about the publicly released pretraining models, \emph{i.e.,} MAE and CLIP, in Table~\ref{tab:more_results}. We can see two conclusions from Table~\ref{tab:more_results}. First, we can see models pretrained on natural images can not naturally increase the performance of human-centric tasks. Second, although CLIP leverages the vision-language pair and more images, it achieves worse performance than MAE which only uses 1.28M images, which illustrates that the languages in the existing datasets may not describe fine-grained information about human bodies and therefore can not be helpful to human-centric tasks.

\section{Visualization of Task-Specific Features}
\begin{figure*}
    \centering
    \includegraphics[width=0.9\linewidth]{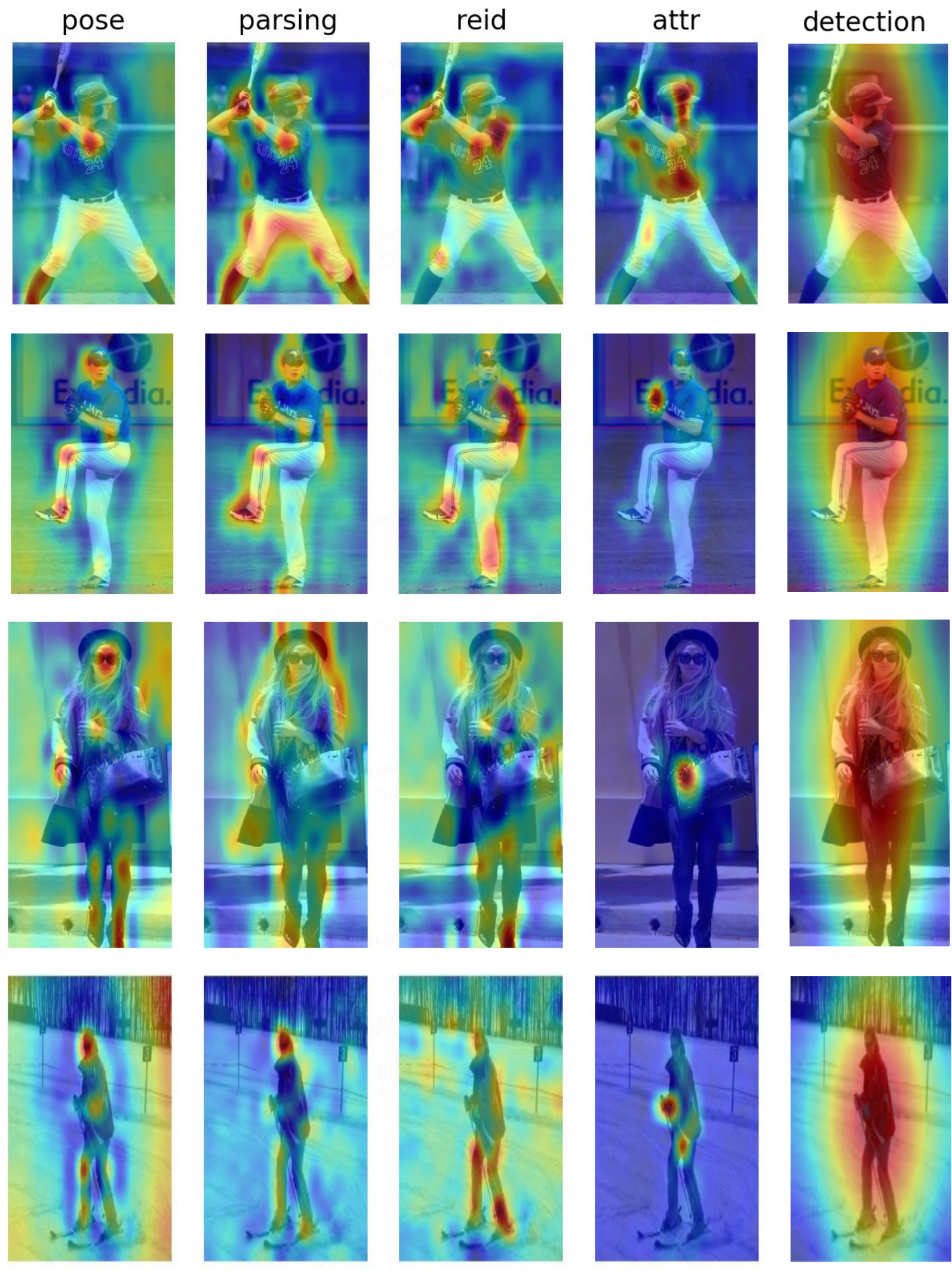}
    \caption{Visualization of features after the task-specific projectors}
    \label{fig:visualization1}
\end{figure*}

\begin{figure*}
    \centering
    \includegraphics[width=0.95\linewidth]{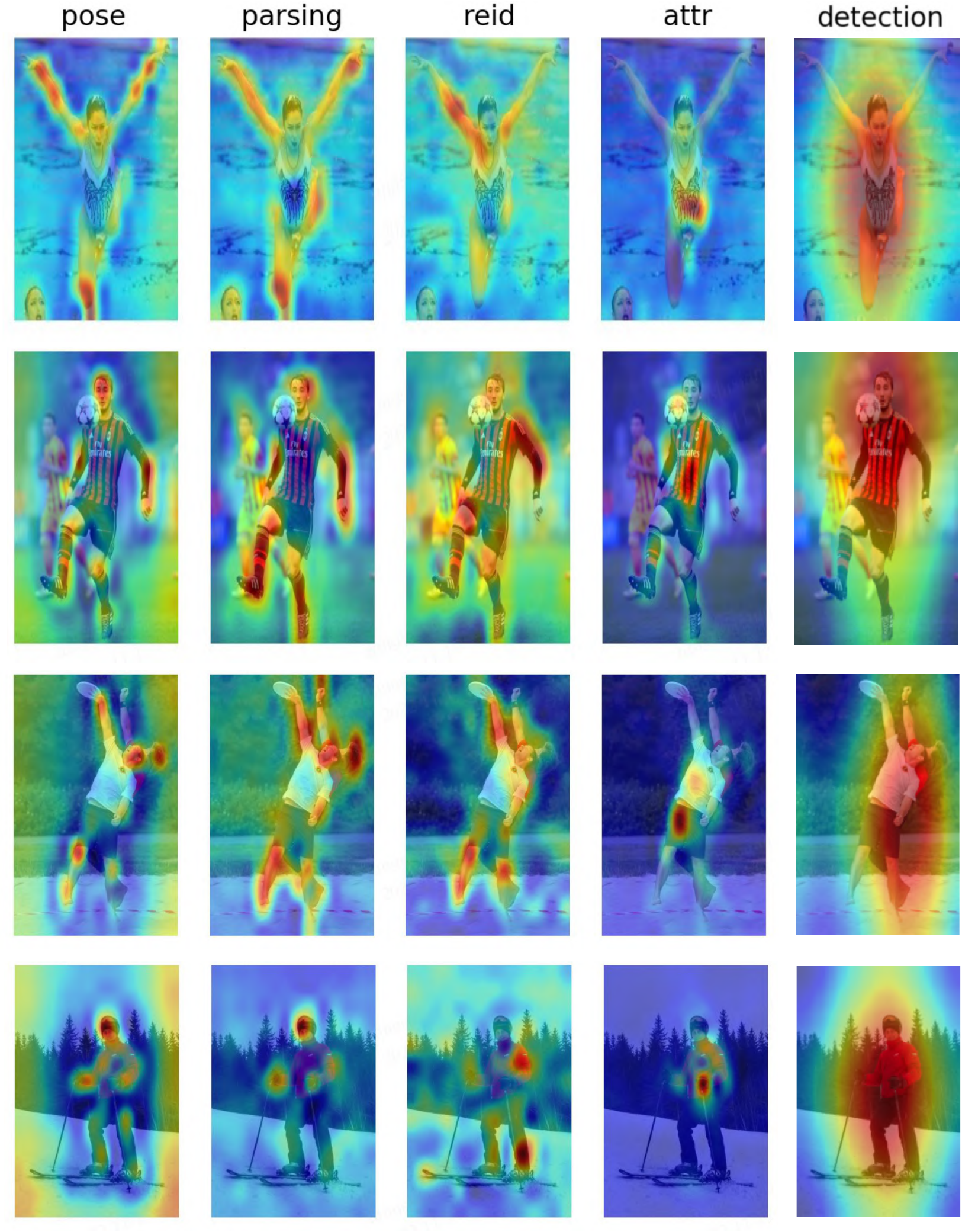}
    \caption{Visualization of features after the task-specific projectors}
    \label{fig:visualization2}
\end{figure*}

\begin{figure*}
    \centering
    \includegraphics[width=0.9\linewidth]{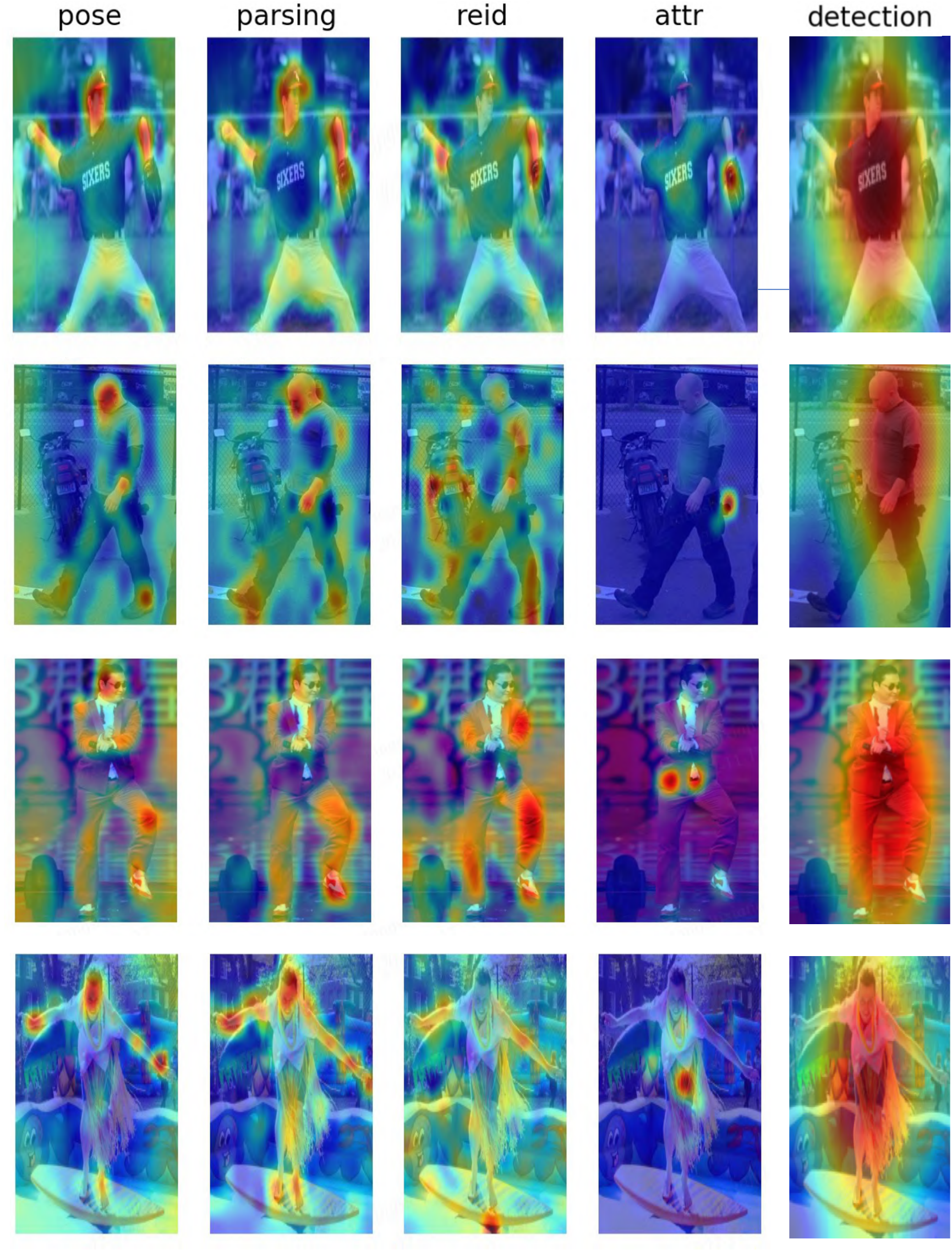}
    \caption{Visualization of features after the task-specific projectors}
    \label{fig:visualization3}
\end{figure*}

To visualize the features attended by the task-specific projectors, we plot the heatmap of L2-normalization of the channels of the attended features. The red color in Figure~\ref{fig:visualization1},~\ref{fig:visualization2},~\ref{fig:visualization3} show the important region, which leads to three conclusions. First, the highlighted regions in the pose estimation and the human parsing locates at the joints of human bodies, which shows that these two tasks are very similar. Second, the heatmap for pedestrian detection includes the whole person, which is consistent with the goal of pedestrian detection to detect all people. Third, for the pedestrian attribute recognition, we can see that the heatmap highlights the attributes, \emph{e.g.,} gloves, bags. These highlighted regions instead of the whole body are also consistent with the goal of pedestrian attribute recognition to recognize attributes.

\section{Detailed Design of PATH}

\subsection{Structure of Attention Module in Projector}
In this section, we describe the structure of attention modules in the projector, which includes a squeeze-and-excitation module and a self-attention module. Specifically, given the feature maps $\mathbf{F}$ and $\mathbf{f}_l \in \mathcal{R}^{C\times H \times W}$ extracted from $l$-th layer, \emph{i.e.,} $\mathbf{F}=(\mathbf{f}_1, \mathbf{f}_2, ..., \mathbf{f}_8)$, the squeeze-and-excitation layer $\mathcal{E}$ transforms the feature $\mathbf{f}_l$ as 
\begin{equation}
    \mathbf{e}_l = \mathcal{E}(\mathbf{f}_l) = \mathcal{F}_{sq}(\frac{1}{H \times W}\sum_{u=1}^H\sum_{v=1}^W f_l(:, u,v)) \odot \mathbf{f}_l,
\end{equation}
where $\mathcal{F}_{sq}$ is the 1-D convolution operation and $\odot$ is the element-wise multiplication of two tensors. Here $ \mathcal{F}_{sq}(\frac{1}{H \times W}\sum_{u=1}^H\sum_{v=1}^W f_l(:, u,v))$ is the channel-wise attention calculated by the squeeze-and-excitation layer.

Next, we feed $\mathbf{s}_l$ into the self-attention module $\mathcal{A}$, which exactly follows~\cite{vaswani2017attention}, which mathmatically can be defined as
\begin{equation}
    \mathbf{p} = \mathcal{A}(\mathbf{e}).
\end{equation}

\subsection{Task Head and Objective Functions}
In this section, we present the task head and the loss designs in Sec. 4.3 in the main text. Given the features $\mathbf{P}^t_j$ after the projector of all image $\mathbf{X}$ for the $j$-th dataset in $t$-th task $\mathcal{D}^t_j$, we compute the losses according to different tasks.

\subsubsection{Person ReID}
\paragraph{Task Head.} Following~\cite{luo2019bag}, the task head of person ReID is a Synchronized BatchNorm~\cite{ioffe2015batch}. Mathematically, the activation $\mathbf{Z}_j^t$ is defined as 
\begin{equation}
    \mathbf{Z}_j^t = \text{BatchNorm}(\mathbf{P}_j^t).
\end{equation}

\paragraph{Objective Function.} We use the triplet loss~\cite{hermans2017defense} and cross-entropy~\cite{zhang2018generalized} to supervise the ReID task. Mathematically,
\begin{equation}
    \mathcal{L}_{\text{reid}} = \sum_{t=1}^T\sum_{j=1}^{N_t} \mathcal{L}_{\text{ce}}(\mathbf{Z}^t_j, \mathbf{Y}^t_j) + \sum_{t=1}^T\sum_{j=1}^{N_t} \mathcal{L}_{\text{triplet}}(\mathbf{Z}^t_j),
\end{equation}
where $\mathcal{L}_{\text{ce}}$ is the cross-entropy loss, $\mathbf{Y}^t_j$ is the labels and $N^t_j$ is the number of images in $\mathcal{D}^t_j$. The triplet loss enlarges the distance between negative pairs and minimizes the distance between positive pairs, which can be mathematically defined as 
\begin{equation}
    \mathcal{L}_{\text{triplet}} = [d_p - d_n + \alpha]_{+},
\end{equation}
where $d_p$ and $d_n$ are feature distances of positive and negative pairs. $\alpha$ is the margin of triple loss, and $[\cdot]$ equals $max(\cdot, 0)$.

\subsubsection{Pose Estimation}
\paragraph{Task Head.} Following~\cite{xu2022vitpose}, the task head is lightweight, processes the features after the task-specific features, and localizes the keypoints. We use the structure of classic decoders in~\cite{xu2022vitpose}, which consists of two deconvolution blocks, each of which contains one deconvolution layer followed by layer normalization and ReLU. Following the common setting of previous methods in pose estimation, each block upsamples the feature maps by 2 times. Mathematically, the activation (the localization heatmaps) can be defined as
\begin{equation}
    \mathbf{Z}^t_j = \text{Conv}_{1\times1}(\text{Deconv}(\text{Deconv}(\mathbf{P}^t_j))),
\end{equation}
where $\mathbf{Z}^t_j \in \mathcal{R}^{\frac{H}{4}\times \frac{W}{4} \times N_k}$, $H$ is the height of the image, $W$ is the width of the image, and $N_k$ is the number of keypoints.
\paragraph{Objective Function.} We leverage the mean square error (MSE) for pose estimation, \emph{i.e.,}
\begin{equation}
    \mathcal{L}_{\text{pose}} = \sum_{t=1}^T\sum_{j=1}^{N_t} \text{MSE}(\mathbf{Z}^t_j, \mathbf{Y}^t_j),
\end{equation}
where $\mathbf{Y}^t_j$ is the ground-truth heatmap of keypoints.

\subsubsection{Human Parsing}
\paragraph{Task Head.} We follow the naive head design of~\cite{zheng2021rethinking} for human parsing. Specifically, the naive head first projects the features after the task-specific projectors to the dimension of category number (\emph{e.g.,} 20 in LIP~\cite{liang2018look}). For this, we adopt a simple 2-layer network with architecture: $1\times1$ Conv+LayerNorm+ReLU+$1\times1$Conv. After that, we simply bilinearly upsample the output to the full image resolution, followed by a classification layer with pixel-wise cross-entropy loss. Mathematically, the task head can be defined as
\begin{equation}
    \mathbf{Z}'^t_j = \text{Conv}_{1\times1}(\text{LayerNorm}(\text{ReLU}(\text{Conv}_{1\times1}(\mathbf{P}^t_j)))),
\end{equation}
\begin{equation}
    \mathbf{Z}^t_j = \text{Upsample}(\mathbf{Z}'^t_j),
\end{equation}
where $\mathbf{Z}^t_j$ is upsampled to the size of input images.

\paragraph{Objective Function.} Following common implementations in~\cite{yuan2110hrformer}, we use the cross-entropy loss to supervise the human parsing. Specifically, the objective function can be defined as
\begin{equation}
    \mathcal{L}_{\text{parsing}} = \sum_{t=1}^T\sum_{j=1}^{N_t} \text{CE}(\mathbf{\mathbf{Z}}^t_j, \mathbf{Y}^t_j),
\end{equation}
where $\mathbf{Y}^t_j \in \mathcal{R}^{H \times W \times N_c}$ is the annotation map whose elements represent the label of the pixel.

\subsubsection{Pedestrian Attribute Recognition}
\paragraph{Task Head.} Following the common implementations in~\cite{li2022label2label}, we only use a fully-connected layer followed by a sigmoid function to project the feature to the activation, which can be mathematically defined as 
\begin{equation}
    \mathbf{Z}^t_j =  \text{Sigmoid}(\text{FC}(\mathbf{Y}^t_j)),
\end{equation}
where $\mathbf{Z}^t_j \in \mathcal{R}^{N \times N_c}$ \text{Fc} is a fully-connected layer, and $N_c$ is the number of attributes in the dataset. 
\paragraph{Objective Function.} Our objective function is the binary cross-entropy loss between the activation and the ground-truth label, which can be mathematically defined as
\begin{equation}
    \mathcal{L}_{\text{attribute}}  = \sum_{t=1}^T\sum_{j=1}^{N_t}\text{BCE}(\mathbf{Z}^t_j, \mathbf{Y}^t_j).
\end{equation}

\subsubsection{Pedestrian Detection}
\paragraph{Task Head.} Following Anchor Detr~\cite{wang2022anchor}, the task head consists of 9 transformer decoder layers, \emph{i.e.,} $\mathcal{D}\!=\!\{\mathcal{D}_1, \mathcal{D}_2, ..., \mathcal{D}_9\}$. The every transformer decoder layer $\mathcal{D}_i$ includes a cross-attention layer, a self-attention layer, and a feedforward network. Therefore, features processed by the decoder $\mathcal{D}_l$ are defined as
\begin{equation}
    \mathbf{P}_l = \mathcal{D}_l(\mathbf{Q}^t_{l-1}, \mathbf{Q}^t_p, \mathbf{P}^t_j, \mathbf{P}_p),
\end{equation}
where $\mathbf{P}_p = proj(\mathcal{A}_\mathbf{P})$, $proj$ is a linear projection, and $\mathcal{A}_\mathbf{P}$ is the coordinates of all tokens in the task-specific feature $\mathbf{P}^t_j$. Similarly, $\mathbf{Q}^t_p = proj(\mathcal{A}_\mathbf{Q})$ refers to a linear projection of the coordinates of learnable anchor points initialized with a uniform distribution following~\cite{wang2022anchor}. 

\paragraph{Objective Function.} Given the features $\mathbf{P}_L$ after the decoder, we use the classification loss, GIoU loss and bounding box loss to supervise the pedestrian detection, \emph{i.e.,}

\begin{equation}
\begin{aligned}
\mathcal{L}_{peddet} = &\lambda_{cls}\mathcal{L}_{cls}(\mathbf{Z}_{cls}, \mathbf{Y}_{cls}) + \lambda_{iou}\mathcal{L}_{iou}(\mathbf{Z}_{bbox}, \mathbf{Y}_{bbox}) \\& + \lambda_{L1}\mathcal{L}_{L1}(\mathbf{Z}_{bbox}, \mathbf{Y}_{bbox}),
\end{aligned}
\end{equation}
where $\mathcal{L}_{cls}$ is the classificatin loss, $\lambda_{iou}$ is the GIoU loss, $\lambda_{L1}$ is L1 loss of the bounding boxes, and $\mathbf{Y}_{cls}$, $\mathbf{Y}_{bbox}$ are annotations of classes and bounding boxes. Here, $\mathbf{Z}_{cls}=f_{cls}(\mathbf{P}_L)$,  $\mathbf{Z}_{bbox}=f_{bbox}(\mathbf{P}_L)$ are linearly projections of $\mathbf{P}_L$, $f_{cls}$ and $f_{bbox}$ are two fully connected layers.

\begin{table}[t]
	\centering
	
	\begin{tabular}{c}
        \hline
        \textbf{Counting Head } \\
           \hline
         Upsample(scale\_factor=2)                     \\
         Conv\{k=(3,3),c=64,s=1\}-BN-ReLU                  \\
          Conv\{k=(3,3),c=32,s=1\}-BN-ReLU                   \\
        Upsample(scale\_factor=2)              \\
         Conv\{k=(3,3),c=16,s=1\}-BN-ReLU                \\
         Conv\{k=(3,3),c=1,s=1\}-ReLU \\
       \hline

	\end{tabular}
	\caption{Detailed architecture of counting head.}
	\label{Table:CP1}
\end{table}

\subsubsection{Crowd Counting}
\paragraph{Task Head.} Table~\ref{Table:CP1} details the configurations of counting head for regressing the density map. In this table, ``Conv\{k(3,3),c64,s1\}-BN-R'' represents the convolutional operation with kernel size of $3 \times 3$, output channels of 64, and stride size of $1$. The ``BN'' and ``ReLU'' mean that the Batch Normalization and ReLU layer are added to this convolutional layer. Specifically, we denote the task head of counting using layers in Table~\ref{Table:CP1} as $\mathcal{H}_{\text{count}}$, \emph{i.e.,}
\begin{equation}
    \mathbf{Z}^t_j = \mathcal{H}_{\text{count}}(\mathbf{P}_j^t).
\end{equation}

\paragraph{Objective Function.} We leverage the MSE between the activation and the ground-truth heatmap to supervise the learning of crowd counting, \emph{i.e.,}
\begin{equation}
    \mathcal{L}_{\text{counting}} = \text{MSE}(\mathbf{Z}^t_j, \mathbf{Y}^t_j),
\end{equation}
where $\mathbf{Y}^t_j$ is the ground-truth heatmap of crowd counting.

\section{Details of Implementations in Pretraining}
During pretraining, we collect in total 39 datasets from person ReID, human parsing, pose estimation, pedestrian attribute recognition, and pedestrian detection. To pretrain the model in a distributed manner, we only train a dataset in each GPU. We pretrain our model using 64 V100-32G GPUs. In the following, we present the task-agnostic parameters and task-specific parameters.

\subsection{Task-agnostic Hyperparameters}
Table~\ref{tab:task_agnostic_pretraining} illustrates the learning hyper-parameters utilized in our pretraining stage. Specifically, we train our model for 80000 iterations in total. During pretraining, we use \textbf{STEP} learning rate decay strategy with a warm-up from $1e^{-7}$ to $5e^{-4}$ during 1500 iterations. we multiply the learning rate $5e^{-4}$ by 0.5, 0.2 and 0.1 at the 40000-th, 60000-th and 76000-th iteration, respectively. The backbone multiplier and the positional multiplier are the ratios of the actual learning rate of the backbone and the positional embedding, respectively, which are all set as 1.0.

\subsection{Task-specific Hyperparameters}
Table~\ref{tab:task_specific} presents the task-specific hyper-parameters of each dataset, including batch size per GPU, the number of GPUs, sample weights, and loss weights. Specifically, the dataset weights are related to sample weights and the number of GPUs:
\begin{equation}
\footnotesize
    \text{loss weight} = \text{sample weight} \times \text{images per GPU} \times \text{number of GPUs}.
\end{equation}
The loss weights of the pose estimation are larger than other tasks because the loss functions used in pose estimation are MSE loss between the predicted heatmaps of keypoints and the heatmaps of the ground truth whose value is very small. For tasks other than pose estimation, the difference between different datasets among different tasks are relatively small.

\begin{table}[t]
  \centering
  \small
  \caption{Detailed description of task-agnostic hyper-parameters in the pretraining stage.}
    \begin{tabular}{clc}
    \hline
    \multirow{9}[2]{*}{lr\_schedule} & type  & Step \\
          & base\_lr & 1.00E-07 \\
          & warmup\_steps & 1500 \\
          & warmup\_lr & 5.00E-04 \\
          & lr\_mults & [0.5, 0.2, 0.1] \\
          & lr\_steps & [40000, 60000, 76000] \\
          & max\_iter & 80000 \\
          & backbone\_multiplier & 1.0 \\
          & pos\_embed\_multiplier & 1.0 \\
    \midrule
    \multirow{8}[2]{*}{optimizer} & type  & Adafactor\_dev \\
          & beta1 & 0.9 \\
          & clip\_beta2 & 0.999 \\
          & clip\_threshold & 0.5 \\
          & decay\_rate & -0.8 \\
          & scale\_parameter & FALSE \\
          & relative\_step & FALSE \\
          & weight\_decay & 0.05 \\
    \midrule
    \multirow{2}[2]{*}{layer\_decay} & num\_layers & 12 \\
          & layer\_decay\_rate & 0.75 \\
    \bottomrule
    \end{tabular}%
  \label{tab:task_agnostic_pretraining}%
\end{table}%

\begin{table*}[t]
\footnotesize
  \centering
  \caption{Detailed Implementation about Task-specific Hyper-parameters}
    \begin{tabular}{cccccc}
    \hline
    Task  & Dataset & Batch Size Per GPU & GPU   & Sample Weight & Loss Weight \\
    \midrule
    \multirow{3}[2]{*}{ReID} & Market1501+MSMT+CUHK03 & 112   & 1     & 5     & 560 \\
          & DGMarket+LaST+PRCC & 96    & 1     & 0.1   & 9.6 \\
          & LUPerson-NL & 192   & 2     & 1     & 384 \\
    \midrule
    \multirow{11}[2]{*}{Pose} & COCO  & 224   & 2     & 8000  & 3584000 \\
          & AIC   & 224   & 2     & 6000  & 2688000 \\
          & PoseTrack & 224   & 1     & 6000  & 1344000 \\
          & JRDB & 224   & 1     & 4000  & 896000 \\
          & MHP   & 96    & 1     & 4000  & 384000 \\
          & UppenAction & 128   & 1     & 4000  & 512000 \\
          & MPI-INF-3DHP  & 128   & 1     & 4000  & 512000 \\
          & Halpe & 64    & 1     & 2000  & 128000 \\
          & 3dhp  & 128   & 1     & 2000  & 256000 \\
          & Human3.6M  & 128   & 1     & 2000  & 256000 \\
          & AIST++  & 128   & 1     & 2000  & 256000 \\
    \midrule
    \multirow{7}[2]{*}{Parsing} & Human3.6M  & 26    & 3     & 20    & 1560 \\
          & LIP   & 18    & 2     & 20    & 720 \\
          & CIHP  & 24    & 2     & 20    & 960 \\
          & VIP   & 16    & 1     & 20    & 320 \\
          & Paper Doll & 24    & 2     & 15    & 720 \\
          & DeepFashion & 32    & 2     & 15    & 960 \\
          & ModaNet & 32    & 1     & 15    & 480 \\
    \midrule
    \multirow{2}[2]{*}{Attribute} & rap2+pa100k & 128   & 1     & 0.1   & 12.8 \\
          & HARDHC+UAV-Human+Parse27k+Market1501-Attribute & 116   & 1     & 0.1   & 11.6 \\
    \midrule
    \multirow{2}[2]{*}{Detection} & CrowdHuman & 2     & 16    & 10    & 320 \\
          & WidePerson+COCO-person+EuroCity Persons+CityPersons & 2     & 16    & 10    & 320 \\
    \bottomrule
    \end{tabular}%
  \label{tab:task_specific}%
\end{table*}%

\subsection{Data Augmentation}
We apply augmentation techniques to human-centric images, ranging from scene images in pedestrian detection to cropped images in person ReID. Here, we list the augmentations below for different tasks.

\paragraph{Person ReID.} For person ReID, we use the same augmentation as in~\cite{luo2019bag}. Specifically, we use the random horizontal flip and random erasing for pretraining. Finally, we resize the input image to size 256$\times$128.

\paragraph{Pose Estimation.} 
For pose estimation, we use the same sugmentation as ViTPose\cite{xu2022vitpose}. Specifically, we use random horizontal flip, half body transform and random scale rotation for pretraining. Finally, we resize the input image to size 256$\times$192.

\paragraph{Human Parsing.}
For human parsing, we use the same augmentation as in~\cite{gong2018instance}. Specifically, we use random crop, random image rotation, and photometric distortion augmentation for pretraining. Particularly, for the human parsing dataset, we also use horizontal random flip augmentation, \emph{e.g.,} Human3.6M, LIP, CIHP, LIP, VIP. Finally, we resize the input image to size 480$\times$480.

\paragraph{Pedestrian Attribute Recognition.}
For pedestrian attribute recognition, we use the same augmentation as in~\cite{li2022label2label}. Specifically, we use random crop and random horizontal flip augmentation for pretraining. Finally, we resize the input image to size 256$\times$192.

\paragraph{Pedestrian Detection.}
For pedestrian detection, we use the same augmentation as in~\cite{zheng2022progressive}. Specifically, we use random horizontal flips and random crop augmentation for pretraining. Finally, we random resize the input image with the longest side bound of 1333 and the shortest side bound of 800 while keeping the height and width ratio.

\paragraph{Crowd Counting.}
For the crowd counting dataset, we use random horizontal flip, random scaling (0.5$\times$ $\sim$ 2$\times$), and random cropping augmentation for pretraining.

\section{Details of Implementations in Evaluation}
For full finetuning, we carefully tune the learning rate $\{1e^{-3}, 5e^{-4}, 1e^{-4}\}$, the weight decay $\{0.05, 0.1, 0.3\}$, drop path rate $\{0.1, 0.3, 0.5\}$, the backbone multiplier $\{0.1, 0.3, 0.5\}$, and report the best performance. We will provide the exact hyperparameters in our released repository after acceptance. For head finetuning and partial finetuning, we specifically set the weight decay as 0, which empirically proved very important in our experiments.

\end{document}

%% file: table_main.tex
\begin{table*}[htbp]
\caption{Experimental results of our PATH and recent state-of-the-art methods (SoTA in the table) on 6 human-centric tasks. The results include 12 \emph{in-dataset evaluations}, 5 \emph{out-of-dataset evaluations} (columns w. gray), \emph{i.e.,} ATR, SenseReID, Caltech, MPII, PETA, and 2 \emph{unseen task evaluations} on the unseen counting task. Following the most commonly-used metrics, for human parsing tasks, we report pACC for ATR, mIoU for others. For ReID, we report Top1 for SenseReID, mAP for others. For pedestrian detection, we report AP for CrowdHuman and Miss Rate for Caltech. For pose estimation, we report End-Point Error for Human3.6M and mAP for others. For attribute, we all report mean accuracy. For Counting, we report Mean Square Error (MSE). $\dag$ indicates that the results are obtained with additional information,  multitask learning, or stronger models. We highlight the \textbf{best} using 
ViT-Base and ViT-Large backbone, respectively. We also highlight these best results in \textcolor{red}{red} if they outperform SoTAs.}

 \begin{minipage}[t]{\textwidth}
  \centering
     \resizebox{\textwidth}{!}{
    \begin{tabular}{cl|cccccccccc}
    \toprule
          &       & \multicolumn{4}{c}{Human Parsing} & \multicolumn{4}{c}{Person ReID} & \multicolumn{2}{c}{Pedestrian  Detection} \\
\cmidrule(r){3-6}    \cmidrule(r){7-10}  \cmidrule(r){11-12}     &       & Human3.6M & LIP   & CIHP  & \cellcolor[rgb]{ .949,  .949,  .949} ATR & Market1501 & MSMT  & CUHK03 & \cellcolor[rgb]{ .949,  .949,  .949} SenseReID & CrowdHuman & \cellcolor[rgb]{ .949,  .949,  .949} Caltech ($\downarrow$) \\
    \midrule
          & SoTA  & 62.5~\cite{hong2022hcmoco}  & 60.3~\cite{liu2022cdgnet}  & 65.6~\cite{liu2022cdgnet}  & \cellcolor[rgb]{ .949,  .949,  .949} 97.4~\cite{liu2022cdgnet}  & 86.8~\cite{he2021transreid}  & 61.0~\cite{he2021transreid}  & 76.4~\cite{jin2020semantics}   & \cellcolor[rgb]{ .949,  .949,  .949} 34.6~\cite{zhao2017spindle}  & 92.1~\cite{zheng2022progressive}  & \cellcolor[rgb]{ .949,  .949,  .949} 46.6\cite{hasan2021generalizable}  \\
          & SoTA $\dag$ & -   & -   & -   & \cellcolor[rgb]{ .949,  .949,  .949} - & 93.0~\cite{zhu2022pass}  & 71.8~\cite{zhu2022pass}  & 77.7~\cite{li2019memory} & \cellcolor[rgb]{ .949,  .949,  .949} - & 92.5~\cite{zheng2022progressive}  & \cellcolor[rgb]{ .949,  .949,  .949} 28.8~\cite{hasan2021generalizable}  \\
    \midrule
    \multirow{6}[4]{*}{ViT-B} & MAE   & 62.0  & 57.2  & 62.9  & \cellcolor[rgb]{ .949,  .949,  .949} 97.4  & 79.2  & 51.5  & 65.8  & \cellcolor[rgb]{ .949,  .949,  .949} 43.8  & 89.6  & \cellcolor[rgb]{ .949,  .949,  .949} 48.1  \\
          & CLIP  & 58.2  & 53.4  &  61.7 & \cellcolor[rgb]{ .949,  .949,  .949} 97.0  & 78.6  & 53.6  & 66.9  & \cellcolor[rgb]{ .949,  .949,  .949} 42.5  & 82.1  & \cellcolor[rgb]{ .949,  .949,  .949} -  \\
\cmidrule{2-12}          & PATH (w/o FT) & 63.9  & 56.3  & 63.9  & \cellcolor[rgb]{ .949,  .949,  .949} - & 88.6  & 66.3  & 77.2  & \cellcolor[rgb]{ .949,  .949,  .949} - & 89.1  & \cellcolor[rgb]{ .949,  .949,  .949} - \\
          & PATH (FT) & \textcolor{red}{\textbf{65.0}}  & \textcolor{red}{\textbf{61.4}}  & \textcolor{red}{\textbf{66.8}}  & \cellcolor[rgb]{ .949,  .949,  .949} \textcolor{red}{\textbf{97.5}}  & \textbf{89.5}  & \textbf{69.1}  & \textcolor{red}{\textbf{82.6}}  & \cellcolor[rgb]{ .949,  .949,  .949} 47.7  & 90.6  & \cellcolor[rgb]{ .949,  .949,  .949} 30.1  \\
          & PATH (Head FT) & 64.1  & 59.9  & 63.3  & \cellcolor[rgb]{ .949,  .949,  .949} 97.1  & -   & -   & -   & \cellcolor[rgb]{ .949,  .949,  .949} - & 90.0  & \cellcolor[rgb]{ .949,  .949,  .949} 31.1 \\
          & PATH (Partial FT) & 63.7  & 60.0  & 63.1  & \cellcolor[rgb]{ .949,  .949,  .949} 97.2  & 88.7  & 66.1  & 79.5  & \cellcolor[rgb]{ .949,  .949,  .949} \textcolor{red}{\textbf{48.2}}  & \textbf{90.9}  & \cellcolor[rgb]{ .949,  .949,  .949} \textcolor{red}{\textbf{28.3}}  \\
    \midrule
    \multirow{2}[2]{*}{ViT-L} & PATH (w/o FT) & 65.0  & \textcolor{red}{\textbf{62.9}}  & 67.1  & \cellcolor[rgb]{ .949,  .949,  .949} - & 91.6  & 72.7  & 83.7  & \cellcolor[rgb]{ .949,  .949,  .949} - & 89.4  & \cellcolor[rgb]{ .949,  .949,  .949} - \\
          & PATH (Partial FT) & \textcolor{red}{\textbf{66.2}}  & 62.6  & \textcolor{red}{\textbf{67.5}}  & \cellcolor[rgb]{ .949,  .949,  .949} \textcolor{red}{\textbf{97.4}}  & \textbf{91.8}  & \textcolor{red}{\textbf{74.7}}  & \textcolor{red}{\textbf{86.0}}  & \cellcolor[rgb]{ .949,  .949,  .949} \textcolor{red}{\textbf{60.0}}  & \textbf{90.8}  & \cellcolor[rgb]{ .949,  .949,  .949} \textcolor{red}{\textbf{28.7}}  \\
    \toprule
	\end{tabular}
	}
  \end{minipage}
  \begin{minipage}[t]{0.996\textwidth}
  \centering
        \resizebox{\textwidth}{!}{
        \begin{tabular}{cl|ccccccc|cc}
    \toprule
          &       & \multicolumn{4}{c}{Pose Estimation} & \multicolumn{3}{c}{Pedestrian Attribute Recognition} & \multicolumn{2}{c}{Counting (unseen task)} \\
\cmidrule(r){3-6}    \cmidrule(r){7-9}  \cmidrule(r){10-11}        &       & COCO  & Human3.6M ($\downarrow$) & AIC   & \cellcolor[rgb]{ .949,  .949,  .949} MPII & PA-100K & RAPv2 & \cellcolor[rgb]{ .949,  .949,  .949} PETA &  ShTech PartA ($\downarrow$) &  ShTech PartB ($\downarrow$) \\
    \midrule
          & SoTA  & 75.8~\cite{xu2022vitpose}  & 7.4~\cite{sun2019deep}   & - & \cellcolor[rgb]{ .949,  .949,  .949} 92.3~\cite{yang2021transpose}  & 83.5~\cite{jia2022learning}  & 81.0~\cite{jia2022learning}  & \cellcolor[rgb]{ .949,  .949,  .949} 87.1~\cite{jia2022learning}  &  94.3~\cite{shu2022crowd}  &  11.0~\cite{shu2022crowd} \\
          & SoTA $\dag$ & 77.1~\cite{xu2022vitpose}  & -   & 32.0~\cite{xu2022vitpose}    & \cellcolor[rgb]{ .949,  .949,  .949} 93.3~\cite{xu2022vitpose} & -   & -   & \cellcolor[rgb]{ .949,  .949,  .949} - &  - &  - \\
    \midrule
    \multirow{6}[4]{*}{ViT-B} & MAE   & 75.8  & 8.2   & 31.8  & \cellcolor[rgb]{ .949,  .949,  .949} 90.1  & 82.3  & 80.8  & \cellcolor[rgb]{ .949,  .949,  .949} 84.6  &  102.1  &  15.5 \\
          & CLIP  & 74.4  & 9.9   & 31.1  & \cellcolor[rgb]{ .949,  .949,  .949} 88.1  & 76.1  & 77.0  & \cellcolor[rgb]{ .949,  .949,  .949} 81.2  &  117.9  & 16.3 \\
\cmidrule{2-11}          & PATH (w/o FT) & 75.0  & 6.9   & 31.1  & \cellcolor[rgb]{ .949,  .949,  .949} -  & -   & -   & \cellcolor[rgb]{ .949,  .949,  .949} - &  - &  - \\
          & PATH (FT) & \textbf{76.3}  & 6.2   & \textcolor{red}{\textbf{35.0}}  & \cellcolor[rgb]{ .949,  .949,  .949} \textcolor{red}{\textbf{93.3}}  & 85.0  & 81.2  & \cellcolor[rgb]{ .949,  .949,  .949} 88.0  &  \textcolor{red}{\textbf{91.7}}  &  \textcolor{red}{\textbf{10.8}} \\
          & PATH (Head FT) & 75.2  & \textcolor{red}{\textbf{6.1}}   & 31.6  & \cellcolor[rgb]{ .949,  .949,  .949} 92.7  & 77.4  & 72.4  & \cellcolor[rgb]{ .949,  .949,  .949} 79.0  &  - &  - \\
          & PATH (Partial FT) & 76.0  & \textcolor{red}{\textbf{6.1}}   & 33.3  & \cellcolor[rgb]{ .949,  .949,  .949} 93.0  & \textcolor{red}{\textbf{86.9}}  & \textcolor{red}{\textbf{83.1}}  & \cellcolor[rgb]{ .949,  .949,  .949} \textcolor{red}{\textbf{89.8}}  &  - &  14.0 \\
\cmidrule{1-11}    \multirow{2}[2]{*}{ViT-L} & PATH (w/o FT) & 74.7  & 7.1   & 25.6  & \cellcolor[rgb]{ .949,  .949,  .949} - & -   & -   & \cellcolor[rgb]{ .949,  .949,  .949} - &  - &  - \\
          & PATH (Partial FT) & \textcolor{red}{\textbf{77.1}}  & \textcolor{red}{\textbf{5.8}}   & \textcolor{red}{\textbf{36.3}}  & \cellcolor[rgb]{ .949,  .949,  .949} \textcolor{red}{\textbf{93.7}}  & \textcolor{red}{\textbf{90.8}}  & \textcolor{red}{\textbf{87.4}}  & \cellcolor[rgb]{ .949,  .949,  .949} \textcolor{red}{\textbf{90.7}}  &  - &  -\\
    \bottomrule
    \end{tabular}
        }
  \end{minipage}
  \label{tab_main_res}
\end{table*}

%% file: table_ablation.tex
\begin{table}[t]
 \footnotesize
  \centering
  \caption{Ablation results.  "A", "S", and "T" respectively denote all shared, specific, and task-shared projector. $\dag$ indicates the results are reported as 1-heavy occluded MR$^{-2}$ for averaging.}
  \resizebox{0.48\textwidth}{!}{

    \begin{tabular}{rc|cccc}
    \toprule
          &       & (a)   & (b)   & (c)  &  (d) \\
    \midrule
    \multirow{2}[2]{*}{}
          & Shared Pos. Embedding &       &       &            & \checkmark \\
          & Projector Share Type & A     & S     & T         & T \\
    \midrule
    \multicolumn{1}{c}{Detection} & Caltech \dag & 60.6  & 57.5  & 60.2   & 60.9  \\
\cmidrule{2-2}    \multicolumn{1}{c}{\multirow{2}{*}{Attribute}} & PA100K & 84.4  & 84.6  & 84.6    & 84.4  \\
          & PETA  & 87.6  & 87.2  & 87.9  & 87.5  \\
\cmidrule{2-2}    \multicolumn{1}{c}{Pose} & MPII  & 92.0  & 92.4  & 92.5   & 92.4  \\
\cmidrule{2-2}    \multicolumn{1}{c}{Parsing} & LIP   & 59.7  & 59.7  & 60.5   & 61.0  \\
\cmidrule{2-2}    \multicolumn{1}{c}{\multirow{2}{*}{ReID}} & Market1501 & 86.2  & 86.5  & 87.1   & 87.6  \\
          & MSMT  & 65.8  & 66.0  & 66.1  & 66.8  \\
    \midrule
          & \textbf{On average} & \textbf{76.6 } & \textbf{76.3 } & \textbf{76.9 } & \textbf{77.2 } \\
    \bottomrule
    \end{tabular}%

}
    \vspace{-2em}
  \label{tab_ablation}%
\end{table}%

%% file: table_ablation_data.tex
\begin{table}[t]
  \centering
  \caption{Comparison with self-supervised pretraining methods on ImageNet and the subset of our HumanBench. $\dag$ indicates the results are repoted as 1-heavy occluded MR$^{-2}$ for averaging.}
  \resizebox{0.48\textwidth}{!}{
  
    \begin{tabular}{cc|cccc}
    \toprule
    \multicolumn{2}{c|}{Pretraining data} & ImageNet-1K & \multicolumn{3}{c}{Our subset} \\
    \cmidrule(r){1-2} \cmidrule(r){3-3} \cmidrule(r){4-6}
    \multicolumn{2}{c|}{Method} & MAE   & MAE   & MOCOv3 & Ours \\
    \midrule
    Detection & Caltech \dag & 51.9  & 58.2  & 57.5  & 60.9  \\
\cmidrule{2-2}    
    \multirow{2}{*}{Attribute} & PA100K & 82.3  & 83.5  & 82.9  & 84.4  \\
          & PETA  & 84.6  & 85.3  & 84.3  & 87.5  \\
\cmidrule{2-2}    Pose  & MPII  & 90.1  & 91.3  & 90.4  & 92.4  \\
\cmidrule{2-2}    Parsing & LIP   & 57.2  & 60.1  & 58.6  & 61.0  \\
\cmidrule{2-2}    
    \multirow{2}{*}{ReID}  & Market1501 & 79.2  & 84.6  & 86.8  & 87.6  \\
          & MSMT  & 51.5  & 64.5  & 67.2  & 66.8  \\
    \midrule
    \multicolumn{2}{c|}{\textbf{On average}} & \textbf{71.0 } & \textbf{75.4 } & \textbf{75.4 } & \textbf{77.2 } \\
    \bottomrule
    \end{tabular}%
    
    }
 \vspace{-1em}
 \label{tab_ablation_data}%
\end{table}%